\documentclass{article}

\usepackage{amsmath}
\usepackage[nonatbib,preprint]{neurips_2024}

\usepackage{apacite}
\usepackage{apacite}
\usepackage[round]{natbib}

\usepackage{doi}

\renewcommand{\doi}[1]{}

\usepackage{hyperref,wasysym}
\usepackage[utf8]{inputenc} % allow utf-8 input
\usepackage[T1]{fontenc}    % use 8-bit T1 fonts
\usepackage{url}            % simple URL typesetting
\usepackage{booktabs}       % professional-quality tables
\usepackage{amsfonts}       % blackboard math symbols
\usepackage{nicefrac}       % compact symbols for 1/2, etc.
\usepackage{microtype}      % microtypography
\usepackage{xcolor}         % colors
\usepackage{indentfirst}
\usepackage{float} 
\usepackage{graphicx}
\DeclareGraphicsExtensions{.pdf,.eps,.tiff,.png,.jpg, .svg}
\usepackage{setspace}
\usepackage{enumerate}
\usepackage{amsmath}
\usepackage{multirow}
\usepackage{subcaption}
\usepackage{listings}
\usepackage{tcolorbox}
\usepackage{etoolbox}

\usepackage{authblk}

\usepackage{amssymb}% http://ctan.org/pkg/amssymb
\usepackage{pifont}% http://ctan.org/pkg/pifont

\title{Enhancing Reasoning Capabilities in SLMs with Reward Guided Dataset Distillation}

\author[]{\textbf{Shreyansh Padarha}\thanks{Corresponding author:\href{mailto:shreyansh.padarha@outlook.com}{shreyansh.padarha@outlook.com} \\The code is available at: \href{https://github.com/shreyansh-2003/AdvDistill}{GitHub Repository}}}

\affil[]{Oxford Internet Institute, University of Oxford, Oxford, UK}

\begin{document}

% \onehalfspacing
\begingroup

\pagenumbering{gobble} 

% \pdfbookmark{Coversheet}{coversheet}
% \input{coversheet}

\endgroup

\clearpage
\pagenumbering{arabic}  % restart page numbering from 1

\newpage

\maketitle

\begin{abstract}
The push to compress and impart the proficiency of Large Language Models (LLMs) into more deployable and efficient Small Language Models (SLMs) has benefited from improvements in knowledge distillation (KD) techniques. These techniques allow a smaller student model to learn from a more capable and larger teacher model's responses. However, distillation often revolves around the student model merely copying the teacher's in-distribution responses, limiting its generalisability. This limitation is amplified on reasoning tasks and can be computationally expensive. In this study, we propose \textbf{AdvDistill}, a reward-guided dataset distillation framework. We utilise multiple generations (responses) from a teacher for each prompt and assign rewards based on rule-based verifiers. These varying and normally distributed rewards serve as weights when training student models. Our methods and their subsequent behavioural analysis demonstrate a significant improvement in student model performance for mathematical and complex reasoning tasks, showcasing the efficacy and benefits of incorporating a rewarding mechanism in dataset distillation processes.
\end{abstract}

%%%%%%%%%%%%%%%%%%%%%%%%%%%%%%%%%%
%%% Introduction 
%%%%%%%%%%%%%%%%%%%%%%%%%%%%%%%%%%
\section{Introduction}
Large Language Models (LLMs) have leapfrogged in capabilities since the incorporation of multi-step reasoning at pre-training and inference time \citep{1,2,3}. These models adhere to scaling laws across text \citep{4}, image \citep{5} and video \citep{6} modality. Scaling laws establish that LLMs follow the power law relationship, dictating the improvement of model performance (decay of loss objective) with increased parameters, training compute and dataset size. In this context, studies such as \citet{7} and \citet{8} counter-intuitively demonstrate the effectiveness of Small Language Models (SLMs), often with less than 10B parameters, performing on par with larger LLMs \citep{9,10}.

SLM related research has picked up momentum since the release of DeepSeek-R1 \citep{11}, a 671B model whose capabilities were distilled into smaller (1.5B and beyond) Qwen family models \citep{12}. While distillation (knowledge \citep{13} and dataset \citep{14}) as a paradigm has been around the machine learning (ML) circuit for quite a while, the successful transfer of complex mathematical reasoning capabilities into smaller models has opened up endless possibilities. Not only are training LLMs GPU intensive, costly and taxing to the environment, they can’t be deployed on smaller resource limited devices. On the other hand, SLMs are suitable for on-device processing and are efficient enough for edge devices \citep{15}. Such on-device deployment enhances safety and trust within end-users regarding the technology \citep{16}.

Smaller models that perform well attribute their abilities to strong pre-training, followed by supervised fine-tuning (SFT) and multi-stage reinforcement learning (RL). Generally, prior to the RL step – post training –  more capable LLMs (teachers)  are compressed into SLMs (students)  through knowledge distillation (KD) (As seen in Gemma2 \citep{23}  and DeepSeekV3 \citep{24}). This distillation of LLMs has been successfully implemented in critical fields of society such as healthcare (drug discovery \citep{17}, clinical decision support \citep{18}, patient interaction \citep {19}) and  education \citep{20,21,22}. But there exist challenges with KD and compression of LLMs. Step-by-step reasoning through Chain of Thought (CoT) and few-shot learning from ICL (In-context learning) examples often involve deeper relationships and abstract reasoning patterns that SLMs fail to replicate \citep{25,26}. This is especially true for sequence-based distillation that uses cross-entropy and top-k tokens \cite{seqKD}, instead of KL-divergence and the entire vocabulary.

Studies have shown that teacher models are often flawed surrogate agents that do not represent the true data distributions within their outputs (logit spaces) \citep{27}. \citet{28} show this through borrowing the `reward hacking' concept from RL and empirically demonstrating how students often perform `teacher hacking’, where the models over optimise to mimic the teacher but in turn stray away from true data distribution. The reason behind student models\footnote{Student models and SLMs would be used synonymously in this study due to the similarities with selected models.} performing poorly on OOD (Out of Distribution) data and teacher hacking stems from LLMs (teachers’) tendencies. These tendencies involve preferentially generating samples with higher likelihood \citep{29}. This results in poor generalizability, as low-probability OOD outputs are ignored by the teacher. To solve this problem, we introduce \textbf{AdvDistill}, a dataset distillation framework that uses high temperature sampling to gather diverse outputs from the teacher. It rewards (relative advantages) these outputs to help the student model distinguish between responses. Our method helps create a loss objective that captures both positive and negative labels, without requiring logit matching. \textbf{AdvDistill}-based models outperform traditional SFT distilled models, especially on OOD tasks.

%%%%%%%%%%%%%%%%%%%%%%%%%%%%%%%%%%
%%% Related Work
%%%%%%%%%%%%%%%%%%%%%%%%%%%%%%%%%%
\section{Related Work}
Primal works of knowledge distillation \citep{13} introduced the concept of soft labels that enforced the probability distribution of a teacher into the student. This later evolved into using attention matrices \citep{30} and output distance-based methods \citep{31}. The loss function when performing supervised KD for language transformers \citep{47} is typically defined as
\begin{equation}
\mathcal{L}_{\text{KD}} = (1-\alpha)\mathcal{L}_{\text{CE}}(y, \sigma(z_S)) + \alpha\mathcal{L}_{\text{KL}}(\sigma(z_T/\tau), \sigma(z_S/\tau))
\end{equation}
where $\mathcal{L}_{\text{CE}}$ is the cross-entropy loss between the student predictions and ground truth, $\mathcal{L}_{\text{KL}}$ is the Kullback-Leibler divergence between the softened probability distributions of the teacher and student models, $\sigma$ denotes the softmax function, $z_T$ and $z_S$ are the logits from the teacher and student models respectively, $\tau$ is a temperature parameter, and $\alpha$ balances the two loss terms.

Recent distillation methods focus more on the structural aspects, such as employing multiple teachers \citep{32,33,34}, implicit curriculum \citep{35, 36} or self-training \citep{37}. Other studies concentrating on reasoning instilled distillation utilise reasoning steps to either decompose them into local adaptations \citep{38} or create weighted token loss functions \citep{39}. Analysing reasoning in SLMs post-distillation often reveals that they over-optimise (often termed as `overthinking' \citep{40}). This was evident in DeepSeek R1 \citep{11} and its models that had their teacher (base model) undergo RL with GRPO (Group Relative Policy Optimization). \citet{41} showed how these R1 distilled models, had longer incorrect responses, a feature believed to have been carry forward from the teacher model when answering difficult questions.

While used interchangeably, dataset distillation is a separate entity within knowledge distillation. Despite the conceptual overlap, dataset distillation focuses on synthesizing a compact representative dataset from a teacher (in our case), whereas KD transfers learned representations from a teacher to student through soft targets (often termed as dark knowledge). KD in LLMs requires the teacher and student models to be from the same family and share the same tokenizers. There are growing methods within KD, such as using reverse KL divergence \citep{minillm}, online distillation and student generated outputs \citep{gkd}, hybrid approaches \citep{distillm, distillm2}, and speculative decoding based methods \citep{speckd}. But this study primarily focuses on distillation carried out through eliciting knowledge from a teacher, and fine-tuning the student on this knowledge.

With the introduction of the concept of `teacher hacking' \citep{28}, on-policy or online distillation has been suggested as a remedy. On-policy distillation, introduced by \citet{42}, uses self-generated student outputs which are guided by the teacher's token-level probabilities. This was recently extended \citep{43} to include speculative decoding to enable the student model to completely replicate intermediate tokens from the teacher, especially during initial warm-up. These methods whilst effective are extremely resource-intensive, expensive and constrained by dataset and sampling budgets. \citet{44} show that all of these methods can be surpassed by using a weaker but cheaper LLM to sample data and fine-tune the smaller model on it. In this study, we implement a similar approach of not using logit matching or traditional on-policy KD. \citet{45} and \citet{46}, similar to our study borrow advantage and reward-based modelling concepts from RL for distillation, but they use multiple teachers and cold-boot students through initial SFT phases.

%%%%%%%%%%%%%%%%%%%%%%%%%%%%%%%%%%
%%% Experimental Setup
%%%%%%%%%%%%%%%%%%%%%%%%%%%%%%%%%%

\section{Experimental Setup}

\begin{figure}[t]
\setlength{\belowcaptionskip}{-5pt} 
    \centering
    \includegraphics[width=\textwidth]{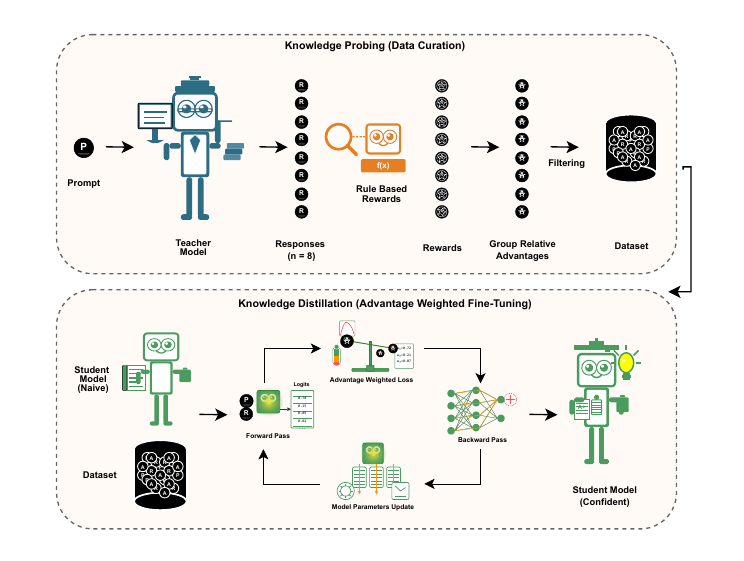}
    \vspace{-2.2em}
    \caption{\textbf{AdvDistill (Group Relative Advantage Distillation) teacher and student knowledge distillation framework.} The teacher model produces 8 different responses for each prompt. The responses are passed through reward functions to generate rewards for each response, which are used for calculating the relative advantage of each response within its group. The student model is trained using the teacher's generated data by using an advantage-guided loss function.}
    \label{figdiag}
    % \vspace{-1.5 em}
\end{figure}

The \texttt{AdvDistill} framework is divided into two stages (\hyperref[figdiag]{Figure 1}). The first stage involves curating a robust dataset from the teacher model. For each prompt (group), 8 responses are generated and put through a rule-based reward function. The generated rewards are used for calculating relative advantages within the response group. A group is accepted into the final dataset if at least one of the responses is correct. The second phase involves fine-tuning a student model with the curated dataset. The fine-tuning uses a custom advantage weighted loss function (\hyperref[sect4]{Section 4}).

%%%%%%%%%%%%%%%%%%%% Models %%%%%%%%%%%%%%%%%%%%
\subsection{Models}
We use models of different sizes from the Qwen 2.5 \citep{12} family. The three base models are Qwen2.5-7B, Qwen2.5-3B and Qwen2.5-1.5B. The 7B model is used as a teacher model for generating responses. For a distillation baseline, the 3B and 1.5B are fine-tuned directly on the best (highest advantage) responses of the teacher. We implement \textbf{AdvDistill} framework on the 1.5B model, and test all the models on test splits and OOD datasets.

%%%%%%%%%%%%%%%%%%%% Dataset %%%%%%%%%%%%%%%%%%%%
\subsection{Dataset}
The datasets used in the study for training the student models (3B, 1.5B) are GSM-8K \citep{50} for mathematics, OPEN-S1 \citep{52} (a filtered version of S1 \citep{54}) for complex mathematical reasoning and MMLU-PRO \citep{51} for multi-task understanding. For testing OOD performance, the models trained on GSM-8K are tested on GSM-PLUS \citep{53} --- a perturbed version of the former --- while models trained on OPEN-S1 are tested on hard difficulty problems from OPEN-RS \citep{52}. All base models are tested with In-context learning (ICL) on the test sets of these datasets.

\begin{table}[H]
\centering
\small
\begin{tabular}{l|c|c}
\toprule
\textbf{Dataset} & \textbf{Train Size} & \textbf{Test Size} \\
\midrule
GSM-8K \citep{50} & 58,232 (7,279 × 8) & 1,319 \\
OPEN-S1 \citep{52} & 19,792 (2,474 × 8) & 553 \\
MMLU-PRO \citep{51} & 55,784 (6,973 × 8) & 2,284 \\
\midrule
GSM-PLUS \citep{53} & – & 2,400 \\
OPEN-RS \citep{52} & – & 850 \\
\bottomrule
\end{tabular}
\vspace{0.5em}
\setlength{\belowcaptionskip}{-5pt}
\caption{\textbf{Dataset sizes used for training (student models) and evaluation (teacher and students).} Datasets with generation-based training from the teacher are shown with expanded response counts.}
\label{datasettable}
\end{table}

\begin{figure}[t]
    
    \centering
    \begin{subfigure}[b]{0.46\textwidth}
        \centering
        \includegraphics[width=\textwidth]{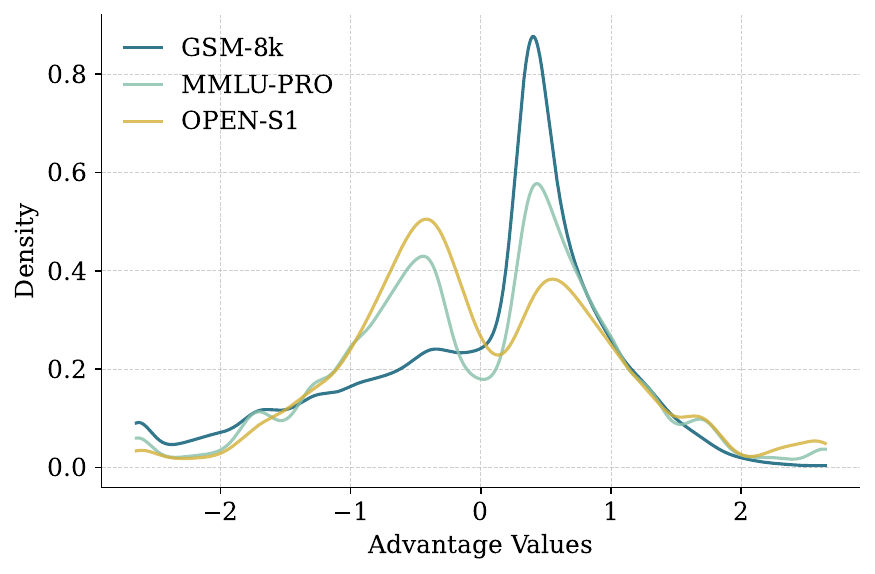}
        \caption{Relative Advantage (Reward) Data Spread}
        \label{advantagesfigure}
    \end{subfigure}
    \hfill
    \begin{subfigure}[b]{0.46\textwidth}
        \centering
        \includegraphics[width=\textwidth]{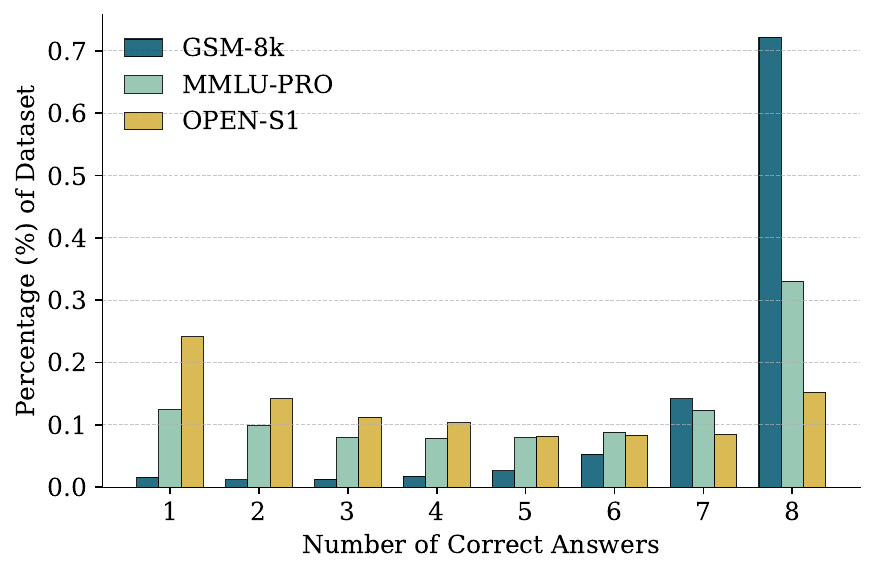}
        \caption{Proportion of Correct Responses}
        \label{correctnumresponsefigure}
    \end{subfigure}
    \setlength{\belowcaptionskip}{-5pt}
    \caption{\textbf{Training dataset distributions and variations.} \textbf{(a)} The spread of advantage values across the datasets calculated through reward functions and group normalization. The values range between -2.5 and 2.5, with multiple peaks. \textbf{(b)} The number of correct answers within each group (prompt and 8 responses). It varies based on teacher's capabilities.}
    \label{fig:mainfig}
\end{figure}

By using high temperature (0.9) response generation from the teacher model, we scale up existing dataset sizes (\hyperref[datasettable]{Table 1}). The curated response advantages are distributed normally with multiple local peaks, serving a wide range of values for weighted loss (\hyperref[advantagesfigure]{Figure 2a}). Due to GSM-8K and its relative ease and outdatedness, the teacher model (Qwen 2.5 7B) produces higher proportion of right answers within each group of 8 responses (\hyperref[correctnumresponsefigure]{Figure 2b}). OPEN-S1 is more challenging for the teacher as its a more complex dataset and requires stronger reasoning abilities. Keeping instances with at least one correct response, allows the student model to see tokens that contribute towards both positive and negative labels.

% \vspace{-1.5em}

%%%%%%%%%%%%%%%%%%%%%%%%%%%%%%%%%%%%%%%%
%% AdvDistill Loss Function and Objective
%%%%%%%%%%%%%%%%%%%%%%%%%%%%%%%%%%%%%%%%
\section{AdvDistill Loss Function and Objective}
\label{sect4}

%%%%%%%%%%%%%%%%%%%% Rule-based Rewards %%%%%%%%%%%%%%%%%%%%
\subsection{Rule-based Rewards}
Neural or model-based rewards typically require an extra policy or oracle model for giving reward signals, adding complexity and computational overhead. Following the Group Relative Policy Optimization (GRPO) framework \citep{48}, we implement rule-based rewards that can be computed deterministically. For a given prompt, the teacher model generates $k$ (8) responses which are then evaluated using a composite rule-based reward function. We handle the inherent bias in GRPO advantage formulation towards longer incorrect responses \citep{41} by using Cosine reward function \citep{49} that scales based on length.

We compute a composite reward for each response $y_j$ as:
\begin{equation}
r_j = w_{\text{cosine}} \times \text{Cosine}_j + w_{\text{format}} \times \text{Format}_j
\end{equation}

where $\text{Cosine}_j \in [-0.5, 1.0]$ is a length-aware reward that varies with correctness and $\text{Format}_j \in \{0,1\}$ is a binary score indicating adherence to formatting. The cosine reward weight ($w_{\text{cosine}}$) is set to 2 and $w_{\text{format}}$ to 1. The Format reward requires thinking steps enclosed in \texttt{<think>} tags, final answers in \texttt{<answer>} tags, and final results in \verb|\boxed{}| notation. The Cosine reward is defined as
\begin{equation}
\text{Cosine}_j = \eta_{\min} + \frac{1}{2}(\eta_{\max} - \eta_{\min})(1 + \cos(\frac{l_j\pi}{L}))
\end{equation}

where $l_j$ is the truncated token length of response $y_j$ (capped at maximum length $L$), and the boundary values $\eta_{\min}$ and $\eta_{\max}$ depend on the correctness of the answer. For correct answers, $\eta_{\min}=0.5, \eta_{\max}=1.0$. And for incorrect answers $\eta_{\min}=-0.5, \eta_{\max}=0.0$. This penalizes lengthy incorrect answers while still rewarding concise correct answers. The value of $L$ is set to the maximum token generation length of the teacher (2048).

\subsubsection{Group Relative Advantages}
For stable training across prompts with varying reward distributions we normalize rewards within each prompt's response group \citep{48}. For each prompt $x$, we compute the relative advantage of each response as
\begin{equation}
A_j = \frac{r_j - \mu}{\sigma + \epsilon}
\end{equation}
where $\mu = \frac{1}{k}\sum_{j=1}^{k}r_j$ is the mean reward, $\sigma = \sqrt{\frac{1}{k}\sum_{j=1}^{k}(r_j - \mu)^2}$ is the standard deviation, and $\epsilon$ is a small constant (set at $10^{-8}$) added for numerical stability. The transformed advantage values represent represent how much better or worse each response is relative to the average response for the same prompt.

%%%%%%%%%%%%%%%%%%%% Loss Function Design %%%%%%%%%%%%%%%%%%%%

\subsection{Loss Function Design} 
The \textbf{AdvDistill} loss function used while training the student model involves two terms and is defined as
\begin{equation}
\mathcal{L}_{\text{AdvDistill}} = \underbrace{\sum_{i=1}^{k} w_{i}\mathcal{L}_{\text{CE}}(y_{i})}_{\text{Advantage-Weighted SFT}} + \lambda_{\text{wrong}}\underbrace{\mathcal{L}_{\text{contrast}}(y_{i})}_{\text{Contrastive Penalty}}
\end{equation}
where $k$ is the number of responses per prompt, $w_{i}$ are advantage-derived weights, and $\lambda_{\text{wrong}}$ controls the strength of the contrastive regularization.

\subsubsection{Advantage-Weighted Supervised Fine-Tuning}
For each response $y_{i}$, we compute a standard cross-entropy loss
\begin{equation} \mathcal{L}_{\text{CE}}(y_{i}) = -\frac{1}{|y_{i}|}\sum_{t=1}^{|y_{i}|} \log \pi_S(y_{i,t}|x, y_{i,<t}) \end{equation}
where $\pi_S(y_{i,t}|x, y_{i,<t})$ is the student model's probability for the correct token $y_{i,t}$ at position (index) $t$.
The weighting scheme $w_{i}$ is calculated using a softmax function over the group relative advantages.

\begin{equation} w_{i} = \frac{\exp(A_{i}/\tau)}{\sum_{i'=1}^{k}\exp(A_{i'}/\tau)} \end{equation}

where $A_{i}$ represents the advantage and $\tau$ is a temperature hyperparameter controlling the weight distribution. Lower $\tau$ values promote responses with higher advantages.

%%%%%%%%%%%%%%%%%% Contrastive Regularization %%%%%%%%%%%%%%%%%

\subsubsection{Contrastive Regularization}
To encourage the student model to assign lower probabilities to tokens in incorrect responses, we use a penalty term. For responses classified as incorrect ($c_{i} = 0$), we apply an additional contrastive term:
\begin{equation} \mathcal{L}_{\text{contrast}}(y{i}) = -\log(1 - \pi_{\text{avg}}(y_{i}|x)) \end{equation}
where $\pi_{\text{avg}}(y_{i}|x)$ is the average token probability:
\begin{equation} \pi_{\text{avg}}(y_{i}|x) = \frac{1}{|y_{i}|}\sum_{t=1}^{|y_{i}|} \pi_S(y_{i,t}|x, y_{i,<t}) \end{equation}
The probabilities are clamped to a maximum value before taking the logarithm and applying gradient clipping with a norm of 1.0. We do this to have numerical stability in the system.

%%%%%%%%%%%%%%%%%%%%%%%%%%%%%%%%%%
%%% Results
%%%%%%%%%%%%%%%%%%%%%%%%%%%%%%%%%%
\section{Results}
\subsection{Performance: Distillation-based Bane or Boon}
\begin{table}[b!]
\centering
\small
\setlength{\tabcolsep}{3pt}
\renewcommand{\arraystretch}{2.5}
\begin{tabular}{l l|c c c c c}
\toprule
\textbf{Model} & \textbf{Method} & \textbf{GSM8K} & \textbf{GSM-PLUS*} & \textbf{OPEN-S1} & \textbf{OPEN-RS*} & \textbf{MMLU-PRO} \\
\midrule
Qwen2.5-7B (Teacher) & BASE$_{\text{ICL}}$ & 88.58\% & 67.83\% & \textcolor{blue}{25.98\%} & \textcolor{blue}{28.28\%} & \textcolor{blue}{37.52\%} \\
\midrule
\multirow{2}{*}{Qwen2.5-3B} & BASE$_{\text{ICL}}$ & 81.22\% & 57.63\% & 20.24\% & 16.89\% & 29.53\% \\
 & SFT$_{\text{Distilled}}$ & 82.08\% & 61.69\% & 21.22\% & 19.97\% & 34.78\% \\
 \midrule
\multirow{3}{*}{Qwen2.5-1.5B} & BASE$_{\text{ICL}}$ & 42.45\% & 30.12\% & 13.52\% & 15.18\% & 15.91\% \\
 & SFT$_{\text{Distilled}}$ & 72.85\% & 51.10\% & 14.68\% & 13.75\% & 30.07\% \\
 & AdvDistill & \textcolor{blue}{91.52\%} & \textcolor{blue}{69.09\%} & 22.77\% & 23.44\% & 23.57\% \\
\bottomrule
\label{tableresultsfinal}
\end{tabular}

\vspace{-0.5em}
\parbox{\textwidth}{* OOD (Out Of Distribution) datasets that the models have not seen in training phase.}

\vspace{1em}
\caption{\textbf{Performance (accuracy) comparison of teacher (7B) and student models (1.5B, 3B) on maths, reasoning and general domain tasks}. Evaluations include: \textbf{(1)} In-context learning with base models, \textbf{(2)} Knowledge Distillation through Supervised Fine-Tuning on Teacher’s strongest outputs, and\textbf{ (3)} Group Relative Advantage Guided Distillation (\texttt{AdvDistill}). The \textcolor{blue}{highest performing model} on simpler mathematical datasets is \texttt{AdvDistill}, while the 7B Teacher excels on more complex reasoning tasks.}
\label{tab:qwen_perf}
\end{table}

The \texttt{AdvDistill} Qwen 2.5 1.5B student model outperforms all base and SFT$_{\text{Distilled}}$ student models (3B and 1.5B) on mathematical and complex reasoning datasets (\hyperref[tableresultsfinal]{Table 1}). We observe a two-fold improvement on test and OOD sets for the 1.5B student over its base model. The model outperforms the approximately 5 times larger teacher model on GSM8K and GSM-PLUS. However, the \texttt{AdvDistill} student struggles on the MMLU-Pro knowledge and multi-task learning dataset, as it fails to improve over its SFT$_{\text{Distilled}}$ counterpart. The 3B Base and SFT$_{\text{Distilled}}$ models outperform it as well. For complex reasoning and multi-task datasets, the teacher is marginally better than the student models. We observe a significant initial delta between the performance capabilities of base models of different sizes. Whilst the performance difference between the 7B and 3B is under 10 percentage points for all objectives, the subsequent performance gap between 3B and 1.5B is more substantial with up to 50 percentage points reduction.

\subsection{Distillation and Optimization Effects}
\subsubsection{Verbosity and Correctness}
\begin{figure}[b]
    \centering
    % Subfigures
    \begin{subfigure}[t]{0.47\textwidth}
        \centering
        \includegraphics[width=\textwidth]{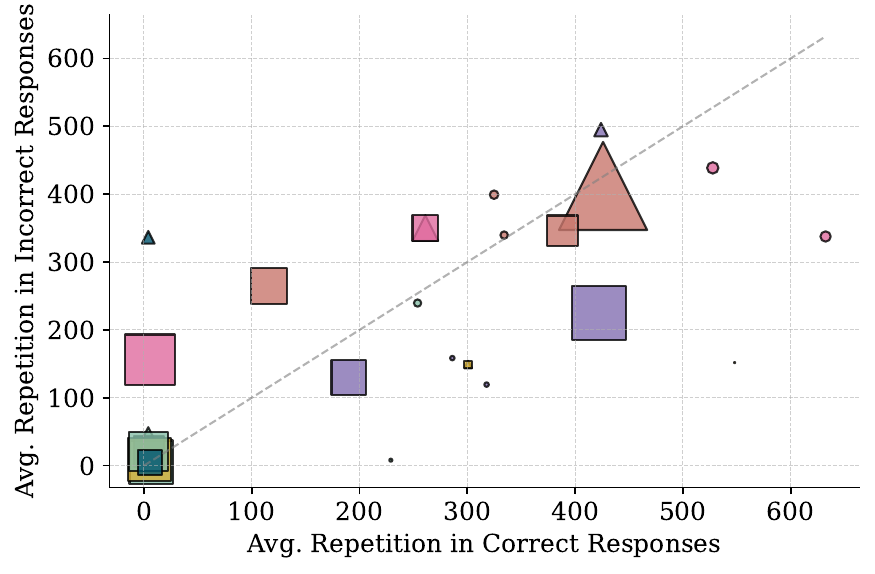}
        \caption{Degenerate Responses}
        \label{figuredegen}
    \end{subfigure}
    \hfill  % This ensures space between the subfigures
    \begin{subfigure}[t]{0.47\textwidth}
        \centering
        \includegraphics[width=\textwidth]{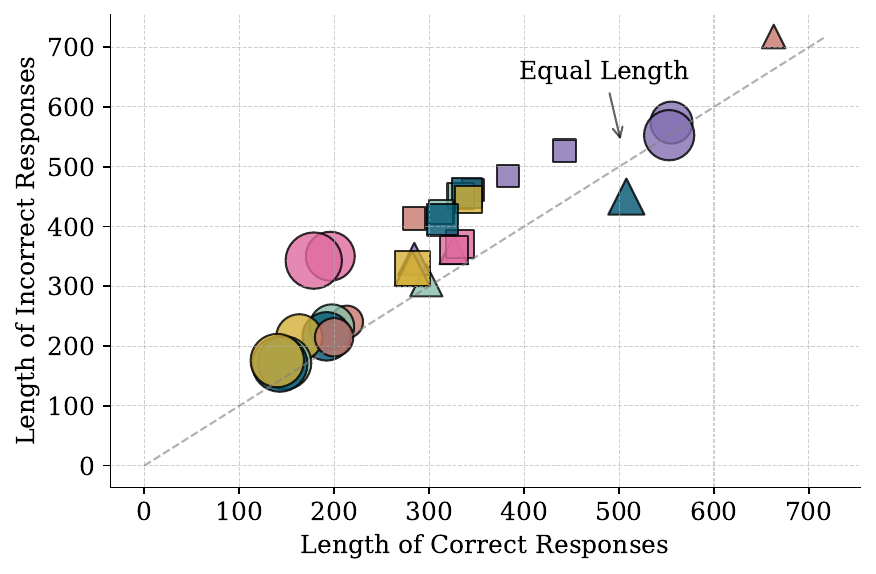}
        \caption{Response Lengths}
        \label{figureresplength}
    \end{subfigure}
    
    % Horizontal model legend (one row)
    \vspace{0.1cm}
    \includegraphics[width=0.9\textwidth]{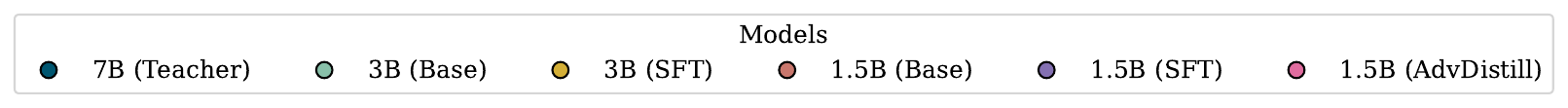}
    
    % Horizontal dataset legend (one row)
    \vspace{0cm}
    \includegraphics[width=0.7\textwidth]{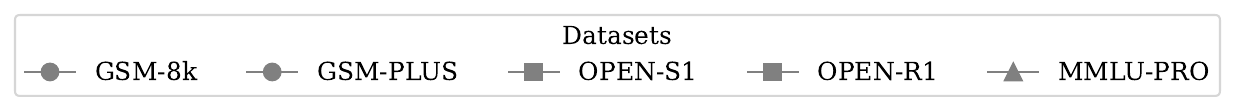}
    \caption{\textbf{Base and distilled model tendencies.} \textbf{(a)} Repetition of tokens in correct and incorrect responses (\hyperref[tabledegenerate]{Table 6}). The \texttt{AdvDistill} student decreases degeneracy (repeating) rate for reasoning tasks but increases it for simpler mathematical datasets. \textbf{(b)} Verbosity of correct and incorrect responses (\hyperref[tableresponselength]{Table 7}). Most models have longer incorrect responses, and some (3B) carry forward this feature from their base to distilled variants.}
    
    \label{bubbleplots}
\end{figure}

The teacher model demonstrates the most balanced response lengths across correct and incorrect responses, with an average token difference of under 50. All model performances for verbosity are extremely task-dependent. The teacher model, whilst having the lowest incorrect response length for mathematical datasets (GSM8K and GSMPLUS), exhibits one of the highest response lengths for incorrect responses on reasoning and multi-task datasets (\hyperref[tableresponselength]{Table 7}, \hyperref[figureresplength]{Figure 3b}) . This pattern is also observed in 3B models. For relative comparison within models, we use the `response verbosity ratio', a metric quantifying the tendency of models to produce longer outputs when generating incorrect answers compared to correct ones (\hyperref[responselengthration]{Figure 6}). This ratio doesn't obey trends across models, but is mostly task-specific. The 1.5B model variants have higher ratios in general and \texttt{AdvDistill} proves ineffective in decreasing this ratio for mathematical datasets. In fact, it has extremely poor ratios for the mathematical datasets (1.92 and 1.79 compared to 1.07 and 1.13 for its base variant). However, with reasoning datasets, we find that the \texttt{AdvDistill} variant performs best overall, showing the lowest difference between correct and incorrect response lengths and an improvement in the verbosity ratio. Specifically with OPEN-R1, we observe the difference (35 tokens) being significantly lower than SFT$_{\text{Distilled}}$ (52 tokens) and Base (129 tokens). Overall, when quantifying the effects of distillation, \texttt{AdvDistill} does not cause universal improvement of response length ratio across all datasets, and SFT$_{\text{Distilled}}$ generally worsens the absolute token count of incorrect responses whilst improving calibration (difference).

\subsubsection{Degeneracy of Tokens}
Degenerate responses include multiple repetitions of the same phrase or words recursively. Within our set of experiments, incorrect responses have higher repetition rates than correct responses (\hyperref[tabledegenerate]{Table 6}, \hyperref[figuredegen]{Figure 3a}). The teacher model across datasets displays the lowest amount of degeneracy. There is a clear pattern with Base and SFT$_{\text{Distilled}}$ models, with the larger sized models performing better. However, with 1.5B, contrary to other models, higher repetition is observed more frequently in correct responses than incorrect ones. The quality improves with the \texttt{AdvDistill} variant, but only for MMLU PRO and OPEN-R1. We find that all Qwen family models experience at least a ten-fold increase in response degeneracy with complex mathematical reasoning datasets.

\subsubsection{Template Adherence: <think> or not to </think>}
\begin{figure}[t!]
    \centering
    \includegraphics[width=0.65\linewidth]{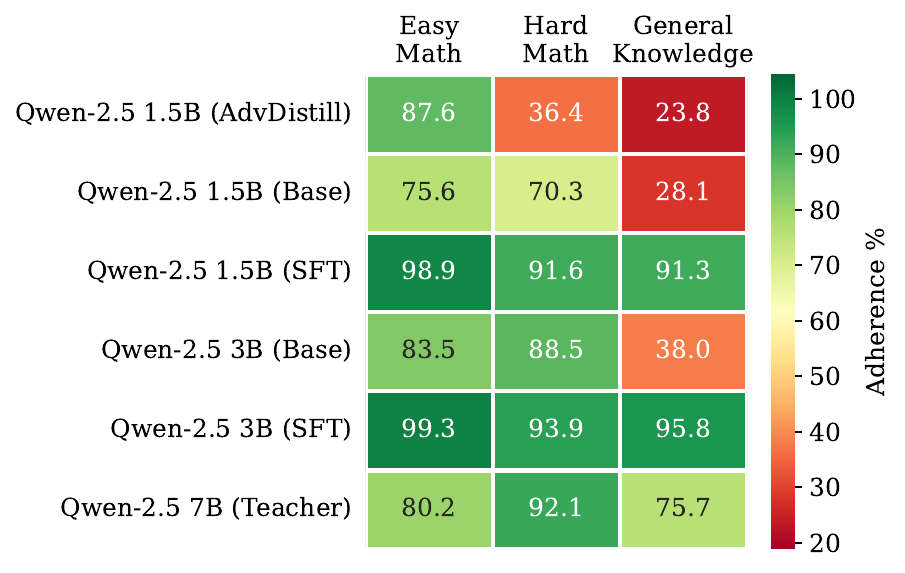}
    \caption{\textbf{Model adherence (\%) to evaluation format.} Both student (1.5B, 3B) and teacher (7B) models find it easier to follow along the evaluation prompts and give responses within the constraints for mathematical and reasoning based datasets.}
    \label{figuretemplateadherence}
\end{figure}

Template adherence refers to models following the requested response format. For this study, within the evaluation prompts, we enforced \verb|<think></think>| tags for reasoning, \verb|<answer></answer>| tags for structured output and \verb|\boxed{}| notation for final option/value. Models generally found it easier to adhere to these notations for mathematical and reasoning-based questions (Figure~\ref{figuretemplateadherence}). SFT$_{\text{Distilled}}$ improved the student models and their response formats, whilst \texttt{AdvDistill} saw the student model significantly deteriorate in template adherence (apart from on the GSM-8K and GSM-PLUS datasets).

%%%%%%%%%%%%%%%%%%%%%%%%%%%%%%%%%%
%%% Discussion
%%%%%%%%%%%%%%%%%%%%%%%%%%%%%%%%%%
\section{Discussion}
\subsection{Domain Specific Performance Variations}
We find significant disparities in the distillation capabilities of \texttt{AdvDistill} across different domains. Whilst mathematical reasoning skills (GSM-8K, GSM-PLUS) transfer remarkably well to the 1.5B model, surpassing even the teacher's performance, we observe limited improvement in multi-task and general knowledge tasks (MMLU-PRO), with SFT$_{\text{Distilled}}$ performing better than our methods for MMLU-PRO. \texttt{AdvDistill} outperforms all models barring the teacher on complex reasoning datasets (OPEN-S1, OPEN-R1). As shown in earlier studies \citep{39}, different reasoning approaches often require different distillation styles. The rule-based approach, through its structured and reward-guided loss, contributes towards the successful transfer of mathematical reasoning. In contrast, multi-task learning for general knowledge (MMLU-PRO) requires memorisation of diverse facts that might not benefit from generalised reward functions. We also examine and speculate that there is a knowledge saturation point within SLMs directly proportional to their size. This is observed through the different carry-down of tendencies from the base model to \texttt{AdvDistill}. Another important consideration is whether advantage-guided fine-tuning (data distillation) on math and reasoning elicits new knowledge capabilities, or reinvigorates new neurons, similar to what \citet{spur} shows with spurious rewards in reinforcement learning. Future work should investigate using different reward-guiding frameworks, such as model-based rewards (LLM as a Judge), varying numbers of responses, and adversarial or counterfactual techniques.

\subsection{Modelling Behaviour and Response Quality}
The \texttt{AdvDistill} framework affects model behaviour beyond performance gains and accuracy metrics. As seen through \hyperref[bubbleplots]{Figure 3}, our approach has a significant impact on both response verbosity and degeneracy rates. For reasoning tasks, we observe \texttt{AdvDistill} successfully reduces token repetition, indicating increased coherence and fluency. However, for simpler mathematical datasets, there is increased degeneracy. Similarly, for response length, \texttt{AdvDistill} produces more balanced lengths between correct and incorrect responses for reasoning tasks, but shows a concerning pattern for simpler mathematical tasks, where incorrect responses are nearly twice as long as correct ones. These patterns suggest that, while \texttt{AdvDistill} increases SLM performance, for certain tasks it induces specific behavioural imbalances. The deterioration in template adherence is particularly noteworthy (\hyperref[figuretemplateadherence]{Figure 4}), as it indicates that advantage-guided training prioritises content over format compliance. Similar to the domain-specific challenges, we find that while our framework improves performance relative to current methods, it requires further refinement and tuning.

\subsection{Practical Trade-offs}
There are several practical considerations with \texttt{AdvDistill}. The computational costs are particularly significant, with \texttt{AdvDistill} requiring approximately 4.5 times more compute (\$108.75) than SFT$_{\text{Distilled}}$ (\$23.75) as shown in \hyperref[computecost]{Table 3}. This substantial resource requirement may limit accessibility for researchers and organisations with constrained computing budgets, potentially creating a divide between those who can and cannot implement such advanced distillation techniques. This represents a trade-off incurred with modern RL-inspired distillation techniques (both on-policy and off-policy).

\begin{table}[H]
\centering
\small
\begin{tabular}{l|ccc|c}
\toprule
\textbf{Modeling} & \textbf{MMLU-PRO} & \textbf{OPEN-S1} & \textbf{GSM-8K} & \textbf{Compute Cost} \\
\midrule
\texttt{AdvDistill} (8 Responses) & 19.5 & 9.0 & 15.0 & \$108.75 (43.5h) \\
\texttt{AdvDistill} (3 Responses) & 11.0 & 4.0 & 7.5 & \$56.25 (22.5h) \\
SFT\textsubscript{Distilled} & 4.0 & 2.5 & 3.0 & \$23.75 (9.5h) \\
\bottomrule
\end{tabular}
\vspace{1em}
\setlength{\belowcaptionskip}{-5pt}
\caption{\textbf{Training duration in GPU hours per dataset and estimated cost (@ \$2.50/hour) across modeling strategies.}}
\label{computecost}
\end{table}

Moreover, for the broader distillation field, the recent implementation of watermarking within proprietary models (e.g., SynthID-Text for Gemini \citep{55}) improves safety by potentially restricting unauthorised distillation of proprietary capabilities. However, it also introduces knowledge inheritance effects on student models that warrant further investigation.

%%%%%%%%%%%%%%%%%%%%%%%%%%%%%%%%%%
%%% Conclusion
%%%%%%%%%%%%%%%%%%%%%%%%%%%%%%%%%%
\section{Conclusion}
With the growing literature on distilling knowledge effectively into SLMs, we propose a novel advantage-guided distillation technique (\texttt{AdvDistill}). The incorporation of group relative advances into distillation demonstrates that a 1.5B parameter model can not only match but sometimes exceed the capabilities of its 7B teacher model, particularly on mathematical reasoning tasks. Our findings reveal nuances within the effectiveness of \texttt{AdvDistill} across different domains, with mathematical reasoning transferring more successfully than general knowledge tasks. The analysis of model behaviour --- including response verbosity, degeneracy patterns, and template adherence --- provides deeper insights into secondary effects of advantage-weighted training beyond performance gains. Future work should explore more efficient implementations of reward-guided distillation, incorporating performance and behavioural checks into teacher response quality and accuracy. The ultimate goal remains developing smaller, more efficient models that retain the reasoning capabilities of their larger counterparts while being deployable in resource-constrained environments. \texttt{AdvDistill} represents a step forward in this direction, contributing to the growing research that brings LLM capabilities to a wider range of devices and applications.

%TC:ignore
\newpage
%%%%%%%%%%%%%%%%%%%%%%%%%%%%%%%%%%
%%% References
%%%%%%%%%%%%%%%%%%%%%%%%%%%%%%%%%%

% \fontsize{10}{1.5}\selectfont

\bibliographystyle{apacite}
\bibliography{refs}

\begin{thebibliography}{}

\bibitem [\protect \citeauthoryear {%
Abdin%
\ \protect \BOthers {.}}{%
Abdin%
\ \protect \BOthers {.}}{%
{\protect \APACyear {2024}}%
}]{%
7}
\APACinsertmetastar {%
7}%
\begin{APACrefauthors}%
Abdin, M.%
, Aneja, J.%
, Awadalla, H.%
, Awadallah, A.%
, Awan, A\BPBI A.%
, Bach, N.%
\BDBL {}Zhou, X.%
\end{APACrefauthors}%
\unskip\
\newblock
\APACrefYearMonthDay{2024}{}{}.
\newblock
\APACrefbtitle {Phi-3 Technical Report: A Highly Capable Language Model Locally on Your Phone.} {Phi-3 technical report: A highly capable language model locally on your phone.}
\PrintBackRefs{\CurrentBib}

\bibitem [\protect \citeauthoryear {%
Agarwal%
\ \protect \BOthers {.}}{%
Agarwal%
\ \protect \BOthers {.}}{%
{\protect \APACyear {2024}}%
{\protect \APACexlab {{\protect \BCnt {1}}}}}]{%
gkd}
\APACinsertmetastar {%
gkd}%
\begin{APACrefauthors}%
Agarwal, R.%
, Vieillard, N.%
, Zhou, Y.%
, Stanczyk, P.%
, Garea, S\BPBI R.%
, Geist, M.%
\BCBL {}\ \BBA {} Bachem, O.%
\end{APACrefauthors}%
\unskip\
\newblock
\APACrefYearMonthDay{2024{\protect \BCnt {1}}}{}{}.
\newblock
{\BBOQ}\APACrefatitle {On-policy distillation of language models: Learning from self-generated mistakes} {On-policy distillation of language models: Learning from self-generated mistakes}.{\BBCQ}
\newblock
\BIn{} \APACrefbtitle {The Twelfth International Conference on Learning Representations.} {The twelfth international conference on learning representations.}
\PrintBackRefs{\CurrentBib}

\bibitem [\protect \citeauthoryear {%
Agarwal%
\ \protect \BOthers {.}}{%
Agarwal%
\ \protect \BOthers {.}}{%
{\protect \APACyear {2024}}%
{\protect \APACexlab {{\protect \BCnt {2}}}}}]{%
42}
\APACinsertmetastar {%
42}%
\begin{APACrefauthors}%
Agarwal, R.%
, Vieillard, N.%
, Zhou, Y.%
, Stanczyk, P.%
, Garea, S\BPBI R.%
, Geist, M.%
\BCBL {}\ \BBA {} Bachem, O.%
\end{APACrefauthors}%
\unskip\
\newblock
\APACrefYearMonthDay{2024{\protect \BCnt {2}}}{}{}.
\newblock
{\BBOQ}\APACrefatitle {On-policy distillation of language models: Learning from self-generated mistakes} {On-policy distillation of language models: Learning from self-generated mistakes}.{\BBCQ}
\newblock
\BIn{} \APACrefbtitle {The Twelfth International Conference on Learning Representations.} {The twelfth international conference on learning representations.}
\PrintBackRefs{\CurrentBib}

\bibitem [\protect \citeauthoryear {%
Baek%
\ \BBA {} Tegmark%
}{%
Baek%
\ \BBA {} Tegmark%
}{%
{\protect \APACyear {2025}}%
}]{%
40}
\APACinsertmetastar {%
40}%
\begin{APACrefauthors}%
Baek, D\BPBI D.%
\BCBT {}\ \BBA {} Tegmark, M.%
\end{APACrefauthors}%
\unskip\
\newblock
\APACrefYearMonthDay{2025}{}{}.
\newblock
{\BBOQ}\APACrefatitle {Towards Understanding Distilled Reasoning Models: A Representational Approach} {Towards understanding distilled reasoning models: A representational approach}.{\BBCQ}
\newblock
\APACjournalVolNumPages{arXiv preprint arXiv:2503.03730}{}{}{}.
\PrintBackRefs{\CurrentBib}

\bibitem [\protect \citeauthoryear {%
Balad{\'o}n%
, Sastre%
, Chiruzzo%
\BCBL {}\ \BBA {} Ros{\'a}%
}{%
Balad{\'o}n%
\ \protect \BOthers {.}}{%
{\protect \APACyear {2023}}%
}]{%
20}
\APACinsertmetastar {%
20}%
\begin{APACrefauthors}%
Balad{\'o}n, A.%
, Sastre, I.%
, Chiruzzo, L.%
\BCBL {}\ \BBA {} Ros{\'a}, A.%
\end{APACrefauthors}%
\unskip\
\newblock
\APACrefYearMonthDay{2023}{}{}.
\newblock
{\BBOQ}\APACrefatitle {RETUYT-InCo at BEA 2023 shared task: Tuning open-source LLMs for generating teacher responses} {Retuyt-inco at bea 2023 shared task: Tuning open-source llms for generating teacher responses}.{\BBCQ}
\newblock
\BIn{} \APACrefbtitle {Proceedings of the 18th Workshop on Innovative Use of NLP for Building Educational Applications (BEA 2023)} {Proceedings of the 18th workshop on innovative use of nlp for building educational applications (bea 2023)}\ (\BPGS\ 756--765).
\PrintBackRefs{\CurrentBib}

\bibitem [\protect \citeauthoryear {%
Bansal%
, Hosseini%
, Agarwal%
, Tran%
\BCBL {}\ \BBA {} Kazemi%
}{%
Bansal%
\ \protect \BOthers {.}}{%
{\protect \APACyear {2024}}%
}]{%
44}
\APACinsertmetastar {%
44}%
\begin{APACrefauthors}%
Bansal, H.%
, Hosseini, A.%
, Agarwal, R.%
, Tran, V\BPBI Q.%
\BCBL {}\ \BBA {} Kazemi, M.%
\end{APACrefauthors}%
\unskip\
\newblock
\APACrefYearMonthDay{2024}{}{}.
\newblock
{\BBOQ}\APACrefatitle {Smaller, weaker, yet better: Training llm reasoners via compute-optimal sampling} {Smaller, weaker, yet better: Training llm reasoners via compute-optimal sampling}.{\BBCQ}
\newblock
\APACjournalVolNumPages{arXiv preprint arXiv:2408.16737}{}{}{}.
\PrintBackRefs{\CurrentBib}

\bibitem [\protect \citeauthoryear {%
Brooks%
\ \protect \BOthers {.}}{%
Brooks%
\ \protect \BOthers {.}}{%
{\protect \APACyear {2024}}%
}]{%
6}
\APACinsertmetastar {%
6}%
\begin{APACrefauthors}%
Brooks, T.%
, Peebles, B.%
, Holmes, C.%
, DePue, W.%
, Guo, Y.%
, Jing, L.%
\BDBL {}others%
\end{APACrefauthors}%
\unskip\
\newblock
\APACrefYearMonthDay{2024}{}{}.
\newblock
{\BBOQ}\APACrefatitle {Video generation models as world simulators. 2024} {Video generation models as world simulators. 2024}.{\BBCQ}
\newblock
\APACjournalVolNumPages{URL https://openai. com/research/video-generation-models-as-world-simulators}{3}{}{1}.
\PrintBackRefs{\CurrentBib}

\bibitem [\protect \citeauthoryear {%
Cobbe%
\ \protect \BOthers {.}}{%
Cobbe%
\ \protect \BOthers {.}}{%
{\protect \APACyear {2021}}%
}]{%
50}
\APACinsertmetastar {%
50}%
\begin{APACrefauthors}%
Cobbe, K.%
, Kosaraju, V.%
, Bavarian, M.%
, Chen, M.%
, Jun, H.%
, Kaiser, L.%
\BDBL {}others%
\end{APACrefauthors}%
\unskip\
\newblock
\APACrefYearMonthDay{2021}{}{}.
\newblock
{\BBOQ}\APACrefatitle {Training verifiers to solve math word problems} {Training verifiers to solve math word problems}.{\BBCQ}
\newblock
\APACjournalVolNumPages{arXiv preprint arXiv:2110.14168}{}{}{}.
\PrintBackRefs{\CurrentBib}

\bibitem [\protect \citeauthoryear {%
Dang%
\ \BBA {} Ngo%
}{%
Dang%
\ \BBA {} Ngo%
}{%
{\protect \APACyear {2025}}%
}]{%
52}
\APACinsertmetastar {%
52}%
\begin{APACrefauthors}%
Dang, Q\BHBI A.%
\BCBT {}\ \BBA {} Ngo, C.%
\end{APACrefauthors}%
\unskip\
\newblock
\APACrefYearMonthDay{2025}{}{}.
\newblock
\APACrefbtitle {Reinforcement Learning for Reasoning in Small LLMs: What Works and What Doesn't.} {Reinforcement learning for reasoning in small llms: What works and what doesn't.}
\PrintBackRefs{\CurrentBib}

\bibitem [\protect \citeauthoryear {%
Dathathri%
\ \protect \BOthers {.}}{%
Dathathri%
\ \protect \BOthers {.}}{%
{\protect \APACyear {2024}}%
}]{%
55}
\APACinsertmetastar {%
55}%
\begin{APACrefauthors}%
Dathathri, S.%
, See, A.%
, Ghaisas, S.%
, Huang, P\BHBI S.%
, McAdam, R.%
, Welbl, J.%
\BDBL {}others%
\end{APACrefauthors}%
\unskip\
\newblock
\APACrefYearMonthDay{2024}{}{}.
\newblock
{\BBOQ}\APACrefatitle {Scalable watermarking for identifying large language model outputs} {Scalable watermarking for identifying large language model outputs}.{\BBCQ}
\newblock
\APACjournalVolNumPages{Nature}{634}{8035}{818--823}.
\PrintBackRefs{\CurrentBib}

\bibitem [\protect \citeauthoryear {%
DeepSeek-AI%
\ \protect \BOthers {.}}{%
DeepSeek-AI%
\ \protect \BOthers {.}}{%
{\protect \APACyear {2025}}%
}]{%
24}
\APACinsertmetastar {%
24}%
\begin{APACrefauthors}%
DeepSeek-AI%
, Liu, A.%
, Feng, B.%
, Xue, B.%
, Wang, B.%
, Wu, B.%
\BDBL {}Pan, Z.%
\end{APACrefauthors}%
\unskip\
\newblock
\APACrefYearMonthDay{2025}{}{}.
\newblock
\APACrefbtitle {DeepSeek-V3 Technical Report.} {Deepseek-v3 technical report.}
\PrintBackRefs{\CurrentBib}

\bibitem [\protect \citeauthoryear {%
Feng%
\ \protect \BOthers {.}}{%
Feng%
\ \protect \BOthers {.}}{%
{\protect \APACyear {2024}}%
{\protect \APACexlab {{\protect \BCnt {1}}}}}]{%
26}
\APACinsertmetastar {%
26}%
\begin{APACrefauthors}%
Feng, K.%
, Li, C.%
, Zhang, X.%
, Zhou, J.%
, Yuan, Y.%
\BCBL {}\ \BBA {} Wang, G.%
\end{APACrefauthors}%
\unskip\
\newblock
\APACrefYearMonthDay{2024{\protect \BCnt {1}}}{}{}.
\newblock
\APACrefbtitle {Keypoint-based Progressive Chain-of-Thought Distillation for LLMs.} {Keypoint-based progressive chain-of-thought distillation for llms.}
\PrintBackRefs{\CurrentBib}

\bibitem [\protect \citeauthoryear {%
Feng%
\ \protect \BOthers {.}}{%
Feng%
\ \protect \BOthers {.}}{%
{\protect \APACyear {2024}}%
{\protect \APACexlab {{\protect \BCnt {2}}}}}]{%
38}
\APACinsertmetastar {%
38}%
\begin{APACrefauthors}%
Feng, K.%
, Li, C.%
, Zhang, X.%
, Zhou, J.%
, Yuan, Y.%
\BCBL {}\ \BBA {} Wang, G.%
\end{APACrefauthors}%
\unskip\
\newblock
\APACrefYearMonthDay{2024{\protect \BCnt {2}}}{}{}.
\newblock
\APACrefbtitle {Keypoint-based Progressive Chain-of-Thought Distillation for LLMs.} {Keypoint-based progressive chain-of-thought distillation for llms.}
\PrintBackRefs{\CurrentBib}

\bibitem [\protect \citeauthoryear {%
Gao%
, Wan%
, Guo%
, Quan%
\BCBL {}\ \BBA {} Wang%
}{%
Gao%
\ \protect \BOthers {.}}{%
{\protect \APACyear {2025}}%
}]{%
45}
\APACinsertmetastar {%
45}%
\begin{APACrefauthors}%
Gao, S.%
, Wan, F.%
, Guo, J.%
, Quan, X.%
\BCBL {}\ \BBA {} Wang, Q.%
\end{APACrefauthors}%
\unskip\
\newblock
\APACrefYearMonthDay{2025}{}{}.
\newblock
{\BBOQ}\APACrefatitle {Advantage-Guided Distillation for Preference Alignment in Small Language Models} {Advantage-guided distillation for preference alignment in small language models}.{\BBCQ}
\newblock
\APACjournalVolNumPages{arXiv preprint arXiv:2502.17927}{}{}{}.
\PrintBackRefs{\CurrentBib}

\bibitem [\protect \citeauthoryear {%
Gu%
, Dong%
, Wei%
\BCBL {}\ \BBA {} Huang%
}{%
Gu%
\ \protect \BOthers {.}}{%
{\protect \APACyear {2023}}%
}]{%
minillm}
\APACinsertmetastar {%
minillm}%
\begin{APACrefauthors}%
Gu, Y.%
, Dong, L.%
, Wei, F.%
\BCBL {}\ \BBA {} Huang, M.%
\end{APACrefauthors}%
\unskip\
\newblock
\APACrefYearMonthDay{2023}{}{}.
\newblock
{\BBOQ}\APACrefatitle {MiniLLM: Knowledge distillation of large language models} {Minillm: Knowledge distillation of large language models}.{\BBCQ}
\newblock
\APACjournalVolNumPages{arXiv preprint arXiv:2306.08543}{}{}{}.
\PrintBackRefs{\CurrentBib}

\bibitem [\protect \citeauthoryear {%
Guan%
\ \protect \BOthers {.}}{%
Guan%
\ \protect \BOthers {.}}{%
{\protect \APACyear {2025}}%
}]{%
8}
\APACinsertmetastar {%
8}%
\begin{APACrefauthors}%
Guan, X.%
, Zhang, L\BPBI L.%
, Liu, Y.%
, Shang, N.%
, Sun, Y.%
, Zhu, Y.%
\BDBL {}Yang, M.%
\end{APACrefauthors}%
\unskip\
\newblock
\APACrefYearMonthDay{2025}{}{}.
\newblock
\APACrefbtitle {rStar-Math: Small LLMs Can Master Math Reasoning with Self-Evolved Deep Thinking.} {rstar-math: Small llms can master math reasoning with self-evolved deep thinking.}
\PrintBackRefs{\CurrentBib}

\bibitem [\protect \citeauthoryear {%
Guo%
\ \protect \BOthers {.}}{%
Guo%
\ \protect \BOthers {.}}{%
{\protect \APACyear {2025}}%
}]{%
11}
\APACinsertmetastar {%
11}%
\begin{APACrefauthors}%
Guo, D.%
, Yang, D.%
, Zhang, H.%
, Song, J.%
, Zhang, R.%
, Xu, R.%
\BDBL {}others%
\end{APACrefauthors}%
\unskip\
\newblock
\APACrefYearMonthDay{2025}{}{}.
\newblock
{\BBOQ}\APACrefatitle {Deepseek-r1: Incentivizing reasoning capability in llms via reinforcement learning} {Deepseek-r1: Incentivizing reasoning capability in llms via reinforcement learning}.{\BBCQ}
\newblock
\APACjournalVolNumPages{arXiv preprint arXiv:2501.12948}{}{}{}.
\PrintBackRefs{\CurrentBib}

\bibitem [\protect \citeauthoryear {%
Hinton%
, Vinyals%
\BCBL {}\ \BBA {} Dean%
}{%
Hinton%
\ \protect \BOthers {.}}{%
{\protect \APACyear {2015}}%
}]{%
13}
\APACinsertmetastar {%
13}%
\begin{APACrefauthors}%
Hinton, G.%
, Vinyals, O.%
\BCBL {}\ \BBA {} Dean, J.%
\end{APACrefauthors}%
\unskip\
\newblock
\APACrefYearMonthDay{2015}{}{}.
\newblock
\APACrefbtitle {Distilling the Knowledge in a Neural Network.} {Distilling the knowledge in a neural network.}
\PrintBackRefs{\CurrentBib}

\bibitem [\protect \citeauthoryear {%
Hoffmann%
\ \protect \BOthers {.}}{%
Hoffmann%
\ \protect \BOthers {.}}{%
{\protect \APACyear {2022}}%
}]{%
4}
\APACinsertmetastar {%
4}%
\begin{APACrefauthors}%
Hoffmann, J.%
, Borgeaud, S.%
, Mensch, A.%
, Buchatskaya, E.%
, Cai, T.%
, Rutherford, E.%
\BDBL {}Sifre, L.%
\end{APACrefauthors}%
\unskip\
\newblock
\APACrefYearMonthDay{2022}{}{}.
\newblock
\APACrefbtitle {Training Compute-Optimal Large Language Models.} {Training compute-optimal large language models.}
\PrintBackRefs{\CurrentBib}

\bibitem [\protect \citeauthoryear {%
Hsieh%
\ \protect \BOthers {.}}{%
Hsieh%
\ \protect \BOthers {.}}{%
{\protect \APACyear {2023}}%
}]{%
25}
\APACinsertmetastar {%
25}%
\begin{APACrefauthors}%
Hsieh, C\BHBI Y.%
, Li, C\BHBI L.%
, Yeh, C\BHBI K.%
, Nakhost, H.%
, Fujii, Y.%
, Ratner, A.%
\BDBL {}Pfister, T.%
\end{APACrefauthors}%
\unskip\
\newblock
\APACrefYearMonthDay{2023}{}{}.
\newblock
\APACrefbtitle {Distilling Step-by-Step! Outperforming Larger Language Models with Less Training Data and Smaller Model Sizes.} {Distilling step-by-step! outperforming larger language models with less training data and smaller model sizes.}
\PrintBackRefs{\CurrentBib}

\bibitem [\protect \citeauthoryear {%
Jiao%
\ \protect \BOthers {.}}{%
Jiao%
\ \protect \BOthers {.}}{%
{\protect \APACyear {2020}}%
}]{%
30}
\APACinsertmetastar {%
30}%
\begin{APACrefauthors}%
Jiao, X.%
, Yin, Y.%
, Shang, L.%
, Jiang, X.%
, Chen, X.%
, Li, L.%
\BDBL {}Liu, Q.%
\end{APACrefauthors}%
\unskip\
\newblock
\APACrefYearMonthDay{2020}{}{}.
\newblock
\APACrefbtitle {TinyBERT: Distilling BERT for Natural Language Understanding.} {Tinybert: Distilling bert for natural language understanding.}
\PrintBackRefs{\CurrentBib}

\bibitem [\protect \citeauthoryear {%
Kang%
\ \protect \BOthers {.}}{%
Kang%
\ \protect \BOthers {.}}{%
{\protect \APACyear {2024}}%
}]{%
2}
\APACinsertmetastar {%
2}%
\begin{APACrefauthors}%
Kang, J.%
, Li, X\BPBI Z.%
, Chen, X.%
, Kazemi, A.%
, Sun, Q.%
, Chen, B.%
\BDBL {}others%
\end{APACrefauthors}%
\unskip\
\newblock
\APACrefYearMonthDay{2024}{}{}.
\newblock
{\BBOQ}\APACrefatitle {Mindstar: Enhancing math reasoning in pre-trained llms at inference time} {Mindstar: Enhancing math reasoning in pre-trained llms at inference time}.{\BBCQ}
\newblock
\APACjournalVolNumPages{arXiv preprint arXiv:2405.16265}{}{}{}.
\PrintBackRefs{\CurrentBib}

\bibitem [\protect \citeauthoryear {%
Kim%
\ \BBA {} Rush%
}{%
Kim%
\ \BBA {} Rush%
}{%
{\protect \APACyear {2016}}%
}]{%
seqKD}
\APACinsertmetastar {%
seqKD}%
\begin{APACrefauthors}%
Kim, Y.%
\BCBT {}\ \BBA {} Rush, A\BPBI M.%
\end{APACrefauthors}%
\unskip\
\newblock
\APACrefYearMonthDay{2016}{}{}.
\newblock
{\BBOQ}\APACrefatitle {Sequence-level knowledge distillation} {Sequence-level knowledge distillation}.{\BBCQ}
\newblock
\BIn{} \APACrefbtitle {Proceedings of the 2016 conference on empirical methods in natural language processing} {Proceedings of the 2016 conference on empirical methods in natural language processing}\ (\BPGS\ 1317--1327).
\PrintBackRefs{\CurrentBib}

\bibitem [\protect \citeauthoryear {%
Ko%
\ \protect \BOthers {.}}{%
Ko%
\ \protect \BOthers {.}}{%
{\protect \APACyear {2025}}%
}]{%
distillm2}
\APACinsertmetastar {%
distillm2}%
\begin{APACrefauthors}%
Ko, J.%
, Chen, T.%
, Kim, S.%
, Ding, T.%
, Liang, L.%
, Zharkov, I.%
\BCBL {}\ \BBA {} Yun, S\BHBI Y.%
\end{APACrefauthors}%
\unskip\
\newblock
\APACrefYearMonthDay{2025}{}{}.
\newblock
{\BBOQ}\APACrefatitle {Distillm-2: A contrastive approach boosts the distillation of llms} {Distillm-2: A contrastive approach boosts the distillation of llms}.{\BBCQ}
\newblock
\APACjournalVolNumPages{arXiv preprint arXiv:2503.07067}{}{}{}.
\PrintBackRefs{\CurrentBib}

\bibitem [\protect \citeauthoryear {%
Ko%
, Kim%
, Chen%
\BCBL {}\ \BBA {} Yun%
}{%
Ko%
\ \protect \BOthers {.}}{%
{\protect \APACyear {2024}}%
}]{%
distillm}
\APACinsertmetastar {%
distillm}%
\begin{APACrefauthors}%
Ko, J.%
, Kim, S.%
, Chen, T.%
\BCBL {}\ \BBA {} Yun, S\BHBI Y.%
\end{APACrefauthors}%
\unskip\
\newblock
\APACrefYearMonthDay{2024}{}{}.
\newblock
{\BBOQ}\APACrefatitle {Distillm: Towards streamlined distillation for large language models} {Distillm: Towards streamlined distillation for large language models}.{\BBCQ}
\newblock
\APACjournalVolNumPages{arXiv preprint arXiv:2402.03898}{}{}{}.
\PrintBackRefs{\CurrentBib}

\bibitem [\protect \citeauthoryear {%
Latif%
, Fang%
, Ma%
\BCBL {}\ \BBA {} Zhai%
}{%
Latif%
\ \protect \BOthers {.}}{%
{\protect \APACyear {2024}}%
}]{%
22}
\APACinsertmetastar {%
22}%
\begin{APACrefauthors}%
Latif, E.%
, Fang, L.%
, Ma, P.%
\BCBL {}\ \BBA {} Zhai, X.%
\end{APACrefauthors}%
\unskip\
\newblock
\APACrefYearMonthDay{2024}{}{}.
\newblock
{\BBOQ}\APACrefatitle {Knowledge distillation of llms for automatic scoring of science assessments} {Knowledge distillation of llms for automatic scoring of science assessments}.{\BBCQ}
\newblock
\BIn{} \APACrefbtitle {International Conference on Artificial Intelligence in Education} {International conference on artificial intelligence in education}\ (\BPGS\ 166--174).
\PrintBackRefs{\CurrentBib}

\bibitem [\protect \citeauthoryear {%
Lewis%
\ \protect \BOthers {.}}{%
Lewis%
\ \protect \BOthers {.}}{%
{\protect \APACyear {2025}}%
}]{%
37}
\APACinsertmetastar {%
37}%
\begin{APACrefauthors}%
Lewis, A.%
, White, M.%
, Liu, J.%
, Koike-Akino, T.%
, Parsons, K.%
\BCBL {}\ \BBA {} Wang, Y.%
\end{APACrefauthors}%
\unskip\
\newblock
\APACrefYearMonthDay{2025}{}{}.
\newblock
{\BBOQ}\APACrefatitle {Winning Big with Small Models: Knowledge Distillation vs. Self-Training for Reducing Hallucination in QA Agents} {Winning big with small models: Knowledge distillation vs. self-training for reducing hallucination in qa agents}.{\BBCQ}
\newblock
\APACjournalVolNumPages{arXiv preprint arXiv:2502.19545}{}{}{}.
\PrintBackRefs{\CurrentBib}

\bibitem [\protect \citeauthoryear {%
Li%
, Cui%
, Zhao%
, Kong%
\BCBL {}\ \BBA {} Bi%
}{%
Li%
\ \protect \BOthers {.}}{%
{\protect \APACyear {2024}}%
}]{%
53}
\APACinsertmetastar {%
53}%
\begin{APACrefauthors}%
Li, Q.%
, Cui, L.%
, Zhao, X.%
, Kong, L.%
\BCBL {}\ \BBA {} Bi, W.%
\end{APACrefauthors}%
\unskip\
\newblock
\APACrefYearMonthDay{2024}{}{}.
\newblock
{\BBOQ}\APACrefatitle {Gsm-plus: A comprehensive benchmark for evaluating the robustness of llms as mathematical problem solvers} {Gsm-plus: A comprehensive benchmark for evaluating the robustness of llms as mathematical problem solvers}.{\BBCQ}
\newblock
\APACjournalVolNumPages{arXiv preprint arXiv:2402.19255}{}{}{}.
\PrintBackRefs{\CurrentBib}

\bibitem [\protect \citeauthoryear {%
C.~Liu%
\ \protect \BOthers {.}}{%
C.~Liu%
\ \protect \BOthers {.}}{%
{\protect \APACyear {2024}}%
}]{%
33}
\APACinsertmetastar {%
33}%
\begin{APACrefauthors}%
Liu, C.%
, Kang, Y.%
, Zhao, F.%
, Kuang, K.%
, Jiang, Z.%
, Sun, C.%
\BCBL {}\ \BBA {} Wu, F.%
\end{APACrefauthors}%
\unskip\
\newblock
\APACrefYearMonthDay{2024}{}{}.
\newblock
{\BBOQ}\APACrefatitle {Evolving knowledge distillation with large language models and active learning} {Evolving knowledge distillation with large language models and active learning}.{\BBCQ}
\newblock
\APACjournalVolNumPages{arXiv preprint arXiv:2403.06414}{}{}{}.
\PrintBackRefs{\CurrentBib}

\bibitem [\protect \citeauthoryear {%
Z.~Liu%
\ \protect \BOthers {.}}{%
Z.~Liu%
\ \protect \BOthers {.}}{%
{\protect \APACyear {2025}}%
}]{%
41}
\APACinsertmetastar {%
41}%
\begin{APACrefauthors}%
Liu, Z.%
, Chen, C.%
, Li, W.%
, Qi, P.%
, Pang, T.%
, Du, C.%
\BDBL {}Lin, M.%
\end{APACrefauthors}%
\unskip\
\newblock
\APACrefYearMonthDay{2025}{}{}.
\newblock
\APACrefbtitle {Understanding R1-Zero-Like Training: A Critical Perspective.} {Understanding r1-zero-like training: A critical perspective.}
\PrintBackRefs{\CurrentBib}

\bibitem [\protect \citeauthoryear {%
Muennighoff%
\ \protect \BOthers {.}}{%
Muennighoff%
\ \protect \BOthers {.}}{%
{\protect \APACyear {2025}}%
}]{%
54}
\APACinsertmetastar {%
54}%
\begin{APACrefauthors}%
Muennighoff, N.%
, Yang, Z.%
, Shi, W.%
, Li, X\BPBI L.%
, Fei-Fei, L.%
, Hajishirzi, H.%
\BDBL {}Hashimoto, T.%
\end{APACrefauthors}%
\unskip\
\newblock
\APACrefYearMonthDay{2025}{}{}.
\newblock
\APACrefbtitle {s1: Simple test-time scaling.} {s1: Simple test-time scaling.}
\PrintBackRefs{\CurrentBib}

\bibitem [\protect \citeauthoryear {%
Nakka%
, Dani%
\BCBL {}\ \BBA {} Saxena%
}{%
Nakka%
\ \protect \BOthers {.}}{%
{\protect \APACyear {2025}}%
}]{%
16}
\APACinsertmetastar {%
16}%
\begin{APACrefauthors}%
Nakka, K.%
, Dani, J.%
\BCBL {}\ \BBA {} Saxena, N.%
\end{APACrefauthors}%
\unskip\
\newblock
\APACrefYearMonthDay{2025}{}{}.
\newblock
\APACrefbtitle {Is On-Device AI Broken and Exploitable? Assessing the Trust and Ethics in Small Language Models.} {Is on-device ai broken and exploitable? assessing the trust and ethics in small language models.}
\PrintBackRefs{\CurrentBib}

\bibitem [\protect \citeauthoryear {%
Niu%
\ \protect \BOthers {.}}{%
Niu%
\ \protect \BOthers {.}}{%
{\protect \APACyear {2025}}%
}]{%
19}
\APACinsertmetastar {%
19}%
\begin{APACrefauthors}%
Niu, S.%
, Ma, J.%
, Lin, H.%
, Bai, L.%
, Wang, Z.%
, Xu, Y.%
\BDBL {}Yang, X.%
\end{APACrefauthors}%
\unskip\
\newblock
\APACrefYearMonthDay{2025}{}{}.
\newblock
\APACrefbtitle {Knowledge-Augmented Multimodal Clinical Rationale Generation for Disease Diagnosis with Small Language Models.} {Knowledge-augmented multimodal clinical rationale generation for disease diagnosis with small language models.}
\PrintBackRefs{\CurrentBib}

\bibitem [\protect \citeauthoryear {%
OpenAI%
\ \protect \BOthers {.}}{%
OpenAI%
\ \protect \BOthers {.}}{%
{\protect \APACyear {2024}}%
}]{%
9}
\APACinsertmetastar {%
9}%
\begin{APACrefauthors}%
OpenAI%
, :%
, Hurst, A.%
, Lerer, A.%
, Goucher, A\BPBI P.%
, Perelman, A.%
\BDBL {}Malkov, Y.%
\end{APACrefauthors}%
\unskip\
\newblock
\APACrefYearMonthDay{2024}{}{}.
\newblock
\APACrefbtitle {GPT-4o System Card.} {Gpt-4o system card.}
\PrintBackRefs{\CurrentBib}

\bibitem [\protect \citeauthoryear {%
Panigrahi%
, Liu%
, Malladi%
, Risteski%
\BCBL {}\ \BBA {} Goel%
}{%
Panigrahi%
\ \protect \BOthers {.}}{%
{\protect \APACyear {2024}}%
}]{%
36}
\APACinsertmetastar {%
36}%
\begin{APACrefauthors}%
Panigrahi, A.%
, Liu, B.%
, Malladi, S.%
, Risteski, A.%
\BCBL {}\ \BBA {} Goel, S.%
\end{APACrefauthors}%
\unskip\
\newblock
\APACrefYearMonthDay{2024}{}{}.
\newblock
{\BBOQ}\APACrefatitle {Progressive distillation induces an implicit curriculum} {Progressive distillation induces an implicit curriculum}.{\BBCQ}
\newblock
\APACjournalVolNumPages{arXiv preprint arXiv:2410.05464}{}{}{}.
\PrintBackRefs{\CurrentBib}

\bibitem [\protect \citeauthoryear {%
Park%
, Kim%
, Lu%
\BCBL {}\ \BBA {} Cho%
}{%
Park%
\ \protect \BOthers {.}}{%
{\protect \APACyear {2019}}%
}]{%
31}
\APACinsertmetastar {%
31}%
\begin{APACrefauthors}%
Park, W.%
, Kim, D.%
, Lu, Y.%
\BCBL {}\ \BBA {} Cho, M.%
\end{APACrefauthors}%
\unskip\
\newblock
\APACrefYearMonthDay{2019}{}{}.
\newblock
{\BBOQ}\APACrefatitle {Relational knowledge distillation} {Relational knowledge distillation}.{\BBCQ}
\newblock
\BIn{} \APACrefbtitle {Proceedings of the IEEE/CVF conference on computer vision and pattern recognition} {Proceedings of the ieee/cvf conference on computer vision and pattern recognition}\ (\BPGS\ 3967--3976).
\PrintBackRefs{\CurrentBib}

\bibitem [\protect \citeauthoryear {%
Peebles%
\ \BBA {} Xie%
}{%
Peebles%
\ \BBA {} Xie%
}{%
{\protect \APACyear {2023}}%
}]{%
5}
\APACinsertmetastar {%
5}%
\begin{APACrefauthors}%
Peebles, W.%
\BCBT {}\ \BBA {} Xie, S.%
\end{APACrefauthors}%
\unskip\
\newblock
\APACrefYearMonthDay{2023}{}{}.
\newblock
{\BBOQ}\APACrefatitle {Scalable diffusion models with transformers} {Scalable diffusion models with transformers}.{\BBCQ}
\newblock
\BIn{} \APACrefbtitle {Proceedings of the IEEE/CVF international conference on computer vision} {Proceedings of the ieee/cvf international conference on computer vision}\ (\BPGS\ 4195--4205).
\PrintBackRefs{\CurrentBib}

\bibitem [\protect \citeauthoryear {%
Qu%
, Yin%
, Yu%
, Wang%
\BCBL {}\ \protect \BOthers {.}}{%
Qu%
\ \protect \BOthers {.}}{%
{\protect \APACyear {2024}}%
}]{%
21}
\APACinsertmetastar {%
21}%
\begin{APACrefauthors}%
Qu, Z.%
, Yin, L.%
, Yu, Z.%
, Wang, W.%
\BCBL {}\ \BOthersPeriod {.}\end{APACrefauthors}%
\unskip\
\newblock
\APACrefYearMonthDay{2024}{}{}.
\newblock
{\BBOQ}\APACrefatitle {CourseGPT-ZH: An Educational Large Language Model Based on Knowledge Distillation Incorporating Prompt Optimization} {Coursegpt-zh: An educational large language model based on knowledge distillation incorporating prompt optimization}.{\BBCQ}
\newblock
\APACjournalVolNumPages{arXiv preprint arXiv:2405.04781}{}{}{}.
\PrintBackRefs{\CurrentBib}

\bibitem [\protect \citeauthoryear {%
Sanh%
, Debut%
, Chaumond%
\BCBL {}\ \BBA {} Wolf%
}{%
Sanh%
\ \protect \BOthers {.}}{%
{\protect \APACyear {2019}}%
}]{%
47}
\APACinsertmetastar {%
47}%
\begin{APACrefauthors}%
Sanh, V.%
, Debut, L.%
, Chaumond, J.%
\BCBL {}\ \BBA {} Wolf, T.%
\end{APACrefauthors}%
\unskip\
\newblock
\APACrefYearMonthDay{2019}{}{}.
\newblock
{\BBOQ}\APACrefatitle {DistilBERT, a distilled version of BERT: smaller, faster, cheaper and lighter} {Distilbert, a distilled version of bert: smaller, faster, cheaper and lighter}.{\BBCQ}
\newblock
\APACjournalVolNumPages{arXiv preprint arXiv:1910.01108}{}{}{}.
\PrintBackRefs{\CurrentBib}

\bibitem [\protect \citeauthoryear {%
R.~Shao%
\ \protect \BOthers {.}}{%
R.~Shao%
\ \protect \BOthers {.}}{%
{\protect \APACyear {2025}}%
}]{%
spur}
\APACinsertmetastar {%
spur}%
\begin{APACrefauthors}%
Shao, R.%
, Li, S\BPBI S.%
, Xin, R.%
, Geng, S.%
, Wang, Y.%
, Oh, S.%
\BDBL {}Zettlemoyer, L.%
\end{APACrefauthors}%
\unskip\
\newblock
\APACrefYearMonthDay{2025}{}{}.
\newblock
\APACrefbtitle {Spurious Rewards: Rethinking Training Signals in RLVR.} {Spurious rewards: Rethinking training signals in rlvr.}
\newblock
\begin{APACrefURL} \url{https://arxiv.org/abs/2506.10947} \end{APACrefURL}
\PrintBackRefs{\CurrentBib}

\bibitem [\protect \citeauthoryear {%
Z.~Shao%
\ \protect \BOthers {.}}{%
Z.~Shao%
\ \protect \BOthers {.}}{%
{\protect \APACyear {2024}}%
}]{%
48}
\APACinsertmetastar {%
48}%
\begin{APACrefauthors}%
Shao, Z.%
, Wang, P.%
, Zhu, Q.%
, Xu, R.%
, Song, J.%
, Bi, X.%
\BDBL {}Guo, D.%
\end{APACrefauthors}%
\unskip\
\newblock
\APACrefYearMonthDay{2024}{}{}.
\newblock
\APACrefbtitle {DeepSeekMath: Pushing the Limits of Mathematical Reasoning in Open Language Models.} {Deepseekmath: Pushing the limits of mathematical reasoning in open language models.}
\PrintBackRefs{\CurrentBib}

\bibitem [\protect \citeauthoryear {%
Shumailov%
\ \protect \BOthers {.}}{%
Shumailov%
\ \protect \BOthers {.}}{%
{\protect \APACyear {2023}}%
}]{%
29}
\APACinsertmetastar {%
29}%
\begin{APACrefauthors}%
Shumailov, I.%
, Shumaylov, Z.%
, Zhao, Y.%
, Gal, Y.%
, Papernot, N.%
\BCBL {}\ \BBA {} Anderson, R.%
\end{APACrefauthors}%
\unskip\
\newblock
\APACrefYearMonthDay{2023}{}{}.
\newblock
{\BBOQ}\APACrefatitle {The curse of recursion: Training on generated data makes models forget} {The curse of recursion: Training on generated data makes models forget}.{\BBCQ}
\newblock
\APACjournalVolNumPages{arXiv preprint arXiv:2305.17493}{}{}{}.
\PrintBackRefs{\CurrentBib}

\bibitem [\protect \citeauthoryear {%
Team%
\ \protect \BOthers {.}}{%
Team%
\ \protect \BOthers {.}}{%
{\protect \APACyear {2025}}%
}]{%
10}
\APACinsertmetastar {%
10}%
\begin{APACrefauthors}%
Team, G.%
, Kamath, A.%
, Ferret, J.%
, Pathak, S.%
, Vieillard, N.%
, Merhej, R.%
\BDBL {}others%
\end{APACrefauthors}%
\unskip\
\newblock
\APACrefYearMonthDay{2025}{}{}.
\newblock
{\BBOQ}\APACrefatitle {Gemma 3 technical report} {Gemma 3 technical report}.{\BBCQ}
\newblock
\APACjournalVolNumPages{arXiv preprint arXiv:2503.19786}{}{}{}.
\PrintBackRefs{\CurrentBib}

\bibitem [\protect \citeauthoryear {%
Team%
\ \protect \BOthers {.}}{%
Team%
\ \protect \BOthers {.}}{%
{\protect \APACyear {2024}}%
}]{%
23}
\APACinsertmetastar {%
23}%
\begin{APACrefauthors}%
Team, G.%
, Riviere, M.%
, Pathak, S.%
, Sessa, P\BPBI G.%
, Hardin, C.%
, Bhupatiraju, S.%
\BDBL {}Andreev, A.%
\end{APACrefauthors}%
\unskip\
\newblock
\APACrefYearMonthDay{2024}{}{}.
\newblock
\APACrefbtitle {Gemma 2: Improving Open Language Models at a Practical Size.} {Gemma 2: Improving open language models at a practical size.}
\PrintBackRefs{\CurrentBib}

\bibitem [\protect \citeauthoryear {%
Tian%
, Han%
, Chen%
, Wang%
\BCBL {}\ \BBA {} Chawla%
}{%
Tian%
\ \protect \BOthers {.}}{%
{\protect \APACyear {2025}}%
}]{%
32}
\APACinsertmetastar {%
32}%
\begin{APACrefauthors}%
Tian, Y.%
, Han, Y.%
, Chen, X.%
, Wang, W.%
\BCBL {}\ \BBA {} Chawla, N\BPBI V.%
\end{APACrefauthors}%
\unskip\
\newblock
\APACrefYearMonthDay{2025}{}{}.
\newblock
{\BBOQ}\APACrefatitle {Beyond answers: Transferring reasoning capabilities to smaller llms using multi-teacher knowledge distillation} {Beyond answers: Transferring reasoning capabilities to smaller llms using multi-teacher knowledge distillation}.{\BBCQ}
\newblock
\BIn{} \APACrefbtitle {Proceedings of the Eighteenth ACM International Conference on Web Search and Data Mining} {Proceedings of the eighteenth acm international conference on web search and data mining}\ (\BPGS\ 251--260).
\PrintBackRefs{\CurrentBib}

\bibitem [\protect \citeauthoryear {%
Tiapkin%
\ \protect \BOthers {.}}{%
Tiapkin%
\ \protect \BOthers {.}}{%
{\protect \APACyear {2025}}%
}]{%
28}
\APACinsertmetastar {%
28}%
\begin{APACrefauthors}%
Tiapkin, D.%
, Calandriello, D.%
, Ferret, J.%
, Perrin, S.%
, Vieillard, N.%
, Ram{\'e}, A.%
\BCBL {}\ \BBA {} Blondel, M.%
\end{APACrefauthors}%
\unskip\
\newblock
\APACrefYearMonthDay{2025}{}{}.
\newblock
{\BBOQ}\APACrefatitle {On Teacher Hacking in Language Model Distillation} {On teacher hacking in language model distillation}.{\BBCQ}
\newblock
\APACjournalVolNumPages{arXiv preprint arXiv:2502.02671}{}{}{}.
\PrintBackRefs{\CurrentBib}

\bibitem [\protect \citeauthoryear {%
Wadhwa%
, Shaib%
, Amir%
\BCBL {}\ \BBA {} Wallace%
}{%
Wadhwa%
\ \protect \BOthers {.}}{%
{\protect \APACyear {2025}}%
}]{%
34}
\APACinsertmetastar {%
34}%
\begin{APACrefauthors}%
Wadhwa, S.%
, Shaib, C.%
, Amir, S.%
\BCBL {}\ \BBA {} Wallace, B\BPBI C.%
\end{APACrefauthors}%
\unskip\
\newblock
\APACrefYearMonthDay{2025}{}{}.
\newblock
{\BBOQ}\APACrefatitle {Who Taught You That? Tracing Teachers in Model Distillation} {Who taught you that? tracing teachers in model distillation}.{\BBCQ}
\newblock
\APACjournalVolNumPages{arXiv preprint arXiv:2502.06659}{}{}{}.
\PrintBackRefs{\CurrentBib}

\bibitem [\protect \citeauthoryear {%
C.~Wang%
\ \protect \BOthers {.}}{%
C.~Wang%
\ \protect \BOthers {.}}{%
{\protect \APACyear {2024}}%
}]{%
3}
\APACinsertmetastar {%
3}%
\begin{APACrefauthors}%
Wang, C.%
, Deng, Y.%
, Lyu, Z.%
, Zeng, L.%
, He, J.%
, Yan, S.%
\BCBL {}\ \BBA {} An, B.%
\end{APACrefauthors}%
\unskip\
\newblock
\APACrefYearMonthDay{2024}{}{}.
\newblock
\APACrefbtitle {Q*: Improving Multi-step Reasoning for LLMs with Deliberative Planning.} {Q*: Improving multi-step reasoning for llms with deliberative planning.}
\PrintBackRefs{\CurrentBib}

\bibitem [\protect \citeauthoryear {%
F.~Wang%
\ \protect \BOthers {.}}{%
F.~Wang%
\ \protect \BOthers {.}}{%
{\protect \APACyear {2024}}%
}]{%
15}
\APACinsertmetastar {%
15}%
\begin{APACrefauthors}%
Wang, F.%
, Zhang, Z.%
, Zhang, X.%
, Wu, Z.%
, Mo, T.%
, Lu, Q.%
\BDBL {}others%
\end{APACrefauthors}%
\unskip\
\newblock
\APACrefYearMonthDay{2024}{}{}.
\newblock
{\BBOQ}\APACrefatitle {A comprehensive survey of small language models in the era of large language models: Techniques, enhancements, applications, collaboration with llms, and trustworthiness} {A comprehensive survey of small language models in the era of large language models: Techniques, enhancements, applications, collaboration with llms, and trustworthiness}.{\BBCQ}
\newblock
\APACjournalVolNumPages{arXiv preprint arXiv:2411.03350}{}{}{}.
\PrintBackRefs{\CurrentBib}

\bibitem [\protect \citeauthoryear {%
T.~Wang%
, Zhu%
, Torralba%
\BCBL {}\ \BBA {} Efros%
}{%
T.~Wang%
\ \protect \BOthers {.}}{%
{\protect \APACyear {2020}}%
}]{%
14}
\APACinsertmetastar {%
14}%
\begin{APACrefauthors}%
Wang, T.%
, Zhu, J\BHBI Y.%
, Torralba, A.%
\BCBL {}\ \BBA {} Efros, A\BPBI A.%
\end{APACrefauthors}%
\unskip\
\newblock
\APACrefYearMonthDay{2020}{}{}.
\newblock
\APACrefbtitle {Dataset Distillation.} {Dataset distillation.}
\PrintBackRefs{\CurrentBib}

\bibitem [\protect \citeauthoryear {%
Y.~Wang%
\ \protect \BOthers {.}}{%
Y.~Wang%
\ \protect \BOthers {.}}{%
{\protect \APACyear {2024}}%
}]{%
51}
\APACinsertmetastar {%
51}%
\begin{APACrefauthors}%
Wang, Y.%
, Ma, X.%
, Zhang, G.%
, Ni, Y.%
, Chandra, A.%
, Guo, S.%
\BDBL {}others%
\end{APACrefauthors}%
\unskip\
\newblock
\APACrefYearMonthDay{2024}{}{}.
\newblock
{\BBOQ}\APACrefatitle {MMLU-Pro: A more robust and challenging multi-task language understanding benchmark} {Mmlu-pro: A more robust and challenging multi-task language understanding benchmark}.{\BBCQ}
\newblock
\BIn{} \APACrefbtitle {The Thirty-eight Conference on Neural Information Processing Systems Datasets and Benchmarks Track.} {The thirty-eight conference on neural information processing systems datasets and benchmarks track.}
\PrintBackRefs{\CurrentBib}

\bibitem [\protect \citeauthoryear {%
Wei%
\ \protect \BOthers {.}}{%
Wei%
\ \protect \BOthers {.}}{%
{\protect \APACyear {2022}}%
}]{%
1}
\APACinsertmetastar {%
1}%
\begin{APACrefauthors}%
Wei, J.%
, Wang, X.%
, Schuurmans, D.%
, Bosma, M.%
, Xia, F.%
, Chi, E.%
\BDBL {}others%
\end{APACrefauthors}%
\unskip\
\newblock
\APACrefYearMonthDay{2022}{}{}.
\newblock
{\BBOQ}\APACrefatitle {Chain-of-thought prompting elicits reasoning in large language models} {Chain-of-thought prompting elicits reasoning in large language models}.{\BBCQ}
\newblock
\APACjournalVolNumPages{Advances in neural information processing systems}{35}{}{24824--24837}.
\PrintBackRefs{\CurrentBib}

\bibitem [\protect \citeauthoryear {%
Wu%
\ \protect \BOthers {.}}{%
Wu%
\ \protect \BOthers {.}}{%
{\protect \APACyear {2024}}%
}]{%
39}
\APACinsertmetastar {%
39}%
\begin{APACrefauthors}%
Wu, Z.%
, Bai, H.%
, Zhang, A.%
, Gu, J.%
, Vydiswaran, V\BPBI V.%
, Jaitly, N.%
\BCBL {}\ \BBA {} Zhang, Y.%
\end{APACrefauthors}%
\unskip\
\newblock
\APACrefYearMonthDay{2024}{}{}.
\newblock
\APACrefbtitle {Divide-or-Conquer? Which Part Should You Distill Your LLM?} {Divide-or-conquer? which part should you distill your llm?}
\PrintBackRefs{\CurrentBib}

\bibitem [\protect \citeauthoryear {%
Xu%
\ \protect \BOthers {.}}{%
Xu%
\ \protect \BOthers {.}}{%
{\protect \APACyear {2024}}%
{\protect \APACexlab {{\protect \BCnt {1}}}}}]{%
speckd}
\APACinsertmetastar {%
speckd}%
\begin{APACrefauthors}%
Xu, W.%
, Han, R.%
, Wang, Z.%
, Le, L\BPBI T.%
, Madeka, D.%
, Li, L.%
\BDBL {}Pfister, T.%
\end{APACrefauthors}%
\unskip\
\newblock
\APACrefYearMonthDay{2024{\protect \BCnt {1}}}{}{}.
\newblock
{\BBOQ}\APACrefatitle {Speculative knowledge distillation: Bridging the teacher-student gap through interleaved sampling} {Speculative knowledge distillation: Bridging the teacher-student gap through interleaved sampling}.{\BBCQ}
\newblock
\APACjournalVolNumPages{arXiv preprint arXiv:2410.11325}{}{}{}.
\PrintBackRefs{\CurrentBib}

\bibitem [\protect \citeauthoryear {%
Xu%
\ \protect \BOthers {.}}{%
Xu%
\ \protect \BOthers {.}}{%
{\protect \APACyear {2024}}%
{\protect \APACexlab {{\protect \BCnt {2}}}}}]{%
43}
\APACinsertmetastar {%
43}%
\begin{APACrefauthors}%
Xu, W.%
, Han, R.%
, Wang, Z.%
, Le, L\BPBI T.%
, Madeka, D.%
, Li, L.%
\BDBL {}Pfister, T.%
\end{APACrefauthors}%
\unskip\
\newblock
\APACrefYearMonthDay{2024{\protect \BCnt {2}}}{}{}.
\newblock
{\BBOQ}\APACrefatitle {Speculative knowledge distillation: Bridging the teacher-student gap through interleaved sampling} {Speculative knowledge distillation: Bridging the teacher-student gap through interleaved sampling}.{\BBCQ}
\newblock
\APACjournalVolNumPages{arXiv preprint arXiv:2410.11325}{}{}{}.
\PrintBackRefs{\CurrentBib}

\bibitem [\protect \citeauthoryear {%
Yang%
\ \protect \BOthers {.}}{%
Yang%
\ \protect \BOthers {.}}{%
{\protect \APACyear {2024}}%
}]{%
12}
\APACinsertmetastar {%
12}%
\begin{APACrefauthors}%
Yang, A.%
, Yang, B.%
, Zhang, B.%
, Hui, B.%
, Zheng, B.%
, Yu, B.%
\BDBL {}others%
\end{APACrefauthors}%
\unskip\
\newblock
\APACrefYearMonthDay{2024}{}{}.
\newblock
{\BBOQ}\APACrefatitle {Qwen2. 5 technical report} {Qwen2. 5 technical report}.{\BBCQ}
\newblock
\APACjournalVolNumPages{arXiv preprint arXiv:2412.15115}{}{}{}.
\PrintBackRefs{\CurrentBib}

\bibitem [\protect \citeauthoryear {%
Yeo%
, Tong%
, Niu%
, Neubig%
\BCBL {}\ \BBA {} Yue%
}{%
Yeo%
\ \protect \BOthers {.}}{%
{\protect \APACyear {2025}}%
}]{%
49}
\APACinsertmetastar {%
49}%
\begin{APACrefauthors}%
Yeo, E.%
, Tong, Y.%
, Niu, M.%
, Neubig, G.%
\BCBL {}\ \BBA {} Yue, X.%
\end{APACrefauthors}%
\unskip\
\newblock
\APACrefYearMonthDay{2025}{}{}.
\newblock
{\BBOQ}\APACrefatitle {Demystifying Long Chain-of-Thought Reasoning in LLMs} {Demystifying long chain-of-thought reasoning in llms}.{\BBCQ}
\newblock
\APACjournalVolNumPages{arXiv preprint arXiv:2502.03373}{}{}{}.
\PrintBackRefs{\CurrentBib}

\bibitem [\protect \citeauthoryear {%
Yue%
, Wang%
, Huang%
\BCBL {}\ \BBA {} Wang%
}{%
Yue%
\ \protect \BOthers {.}}{%
{\protect \APACyear {2024}}%
}]{%
35}
\APACinsertmetastar {%
35}%
\begin{APACrefauthors}%
Yue, Y.%
, Wang, C.%
, Huang, J.%
\BCBL {}\ \BBA {} Wang, P.%
\end{APACrefauthors}%
\unskip\
\newblock
\APACrefYearMonthDay{2024}{}{}.
\newblock
\APACrefbtitle {Distilling Instruction-following Abilities of Large Language Models with Task-aware Curriculum Planning.} {Distilling instruction-following abilities of large language models with task-aware curriculum planning.}
\PrintBackRefs{\CurrentBib}

\bibitem [\protect \citeauthoryear {%
K.~Zhang%
\ \protect \BOthers {.}}{%
K.~Zhang%
\ \protect \BOthers {.}}{%
{\protect \APACyear {2025}}%
{\protect \APACexlab {{\protect \BCnt {1}}}}}]{%
17}
\APACinsertmetastar {%
17}%
\begin{APACrefauthors}%
Zhang, K.%
, Zhu, R.%
, Ma, S.%
, Xiong, J.%
, Kim, Y.%
, Murai, F.%
\BCBL {}\ \BBA {} Liu, X.%
\end{APACrefauthors}%
\unskip\
\newblock
\APACrefYearMonthDay{2025{\protect \BCnt {1}}}{}{}.
\newblock
\APACrefbtitle {KEDRec-LM: A Knowledge-distilled Explainable Drug Recommendation Large Language Model.} {Kedrec-lm: A knowledge-distilled explainable drug recommendation large language model.}
\PrintBackRefs{\CurrentBib}

\bibitem [\protect \citeauthoryear {%
K.~Zhang%
\ \protect \BOthers {.}}{%
K.~Zhang%
\ \protect \BOthers {.}}{%
{\protect \APACyear {2025}}%
{\protect \APACexlab {{\protect \BCnt {2}}}}}]{%
18}
\APACinsertmetastar {%
18}%
\begin{APACrefauthors}%
Zhang, K.%
, Zhu, R.%
, Ma, S.%
, Xiong, J.%
, Kim, Y.%
, Murai, F.%
\BCBL {}\ \BBA {} Liu, X.%
\end{APACrefauthors}%
\unskip\
\newblock
\APACrefYearMonthDay{2025{\protect \BCnt {2}}}{}{}.
\newblock
\APACrefbtitle {KEDRec-LM: A Knowledge-distilled Explainable Drug Recommendation Large Language Model.} {Kedrec-lm: A knowledge-distilled explainable drug recommendation large language model.}
\PrintBackRefs{\CurrentBib}

\bibitem [\protect \citeauthoryear {%
R.~Zhang%
\ \protect \BOthers {.}}{%
R.~Zhang%
\ \protect \BOthers {.}}{%
{\protect \APACyear {2023}}%
}]{%
27}
\APACinsertmetastar {%
27}%
\begin{APACrefauthors}%
Zhang, R.%
, Shen, J.%
, Liu, T.%
, Liu, J.%
, Bendersky, M.%
, Najork, M.%
\BCBL {}\ \BBA {} Zhang, C.%
\end{APACrefauthors}%
\unskip\
\newblock
\APACrefYearMonthDay{2023}{}{}.
\newblock
\APACrefbtitle {Do Not Blindly Imitate the Teacher: Using Perturbed Loss for Knowledge Distillation.} {Do not blindly imitate the teacher: Using perturbed loss for knowledge distillation.}
\PrintBackRefs{\CurrentBib}

\bibitem [\protect \citeauthoryear {%
Y.~Zhang%
\ \protect \BOthers {.}}{%
Y.~Zhang%
\ \protect \BOthers {.}}{%
{\protect \APACyear {2025}}%
}]{%
46}
\APACinsertmetastar {%
46}%
\begin{APACrefauthors}%
Zhang, Y.%
, Wang, L.%
, Fang, M.%
, Du, Y.%
, Huang, C.%
, Wang, J.%
\BDBL {}others%
\end{APACrefauthors}%
\unskip\
\newblock
\APACrefYearMonthDay{2025}{}{}.
\newblock
{\BBOQ}\APACrefatitle {Distill Not Only Data but Also Rewards: Can Smaller Language Models Surpass Larger Ones?} {Distill not only data but also rewards: Can smaller language models surpass larger ones?}{\BBCQ}
\newblock
\APACjournalVolNumPages{arXiv preprint arXiv:2502.19557}{}{}{}.
\PrintBackRefs{\CurrentBib}

\end{thebibliography}
% \addbibresource{refs.bib}

%%%%%%%%%%%%%%%%%%%%%%%%%%%%%%%%%%
%%% Apendix
%%%%%%%%%%%%%%%%%%%%%%%%%%%%%%%%%%

\newpage

\appendix

\section*{Appendix} 

\section{Model Outputs}
% Vertical layout with full-width boxes
\begin{figure}[H]
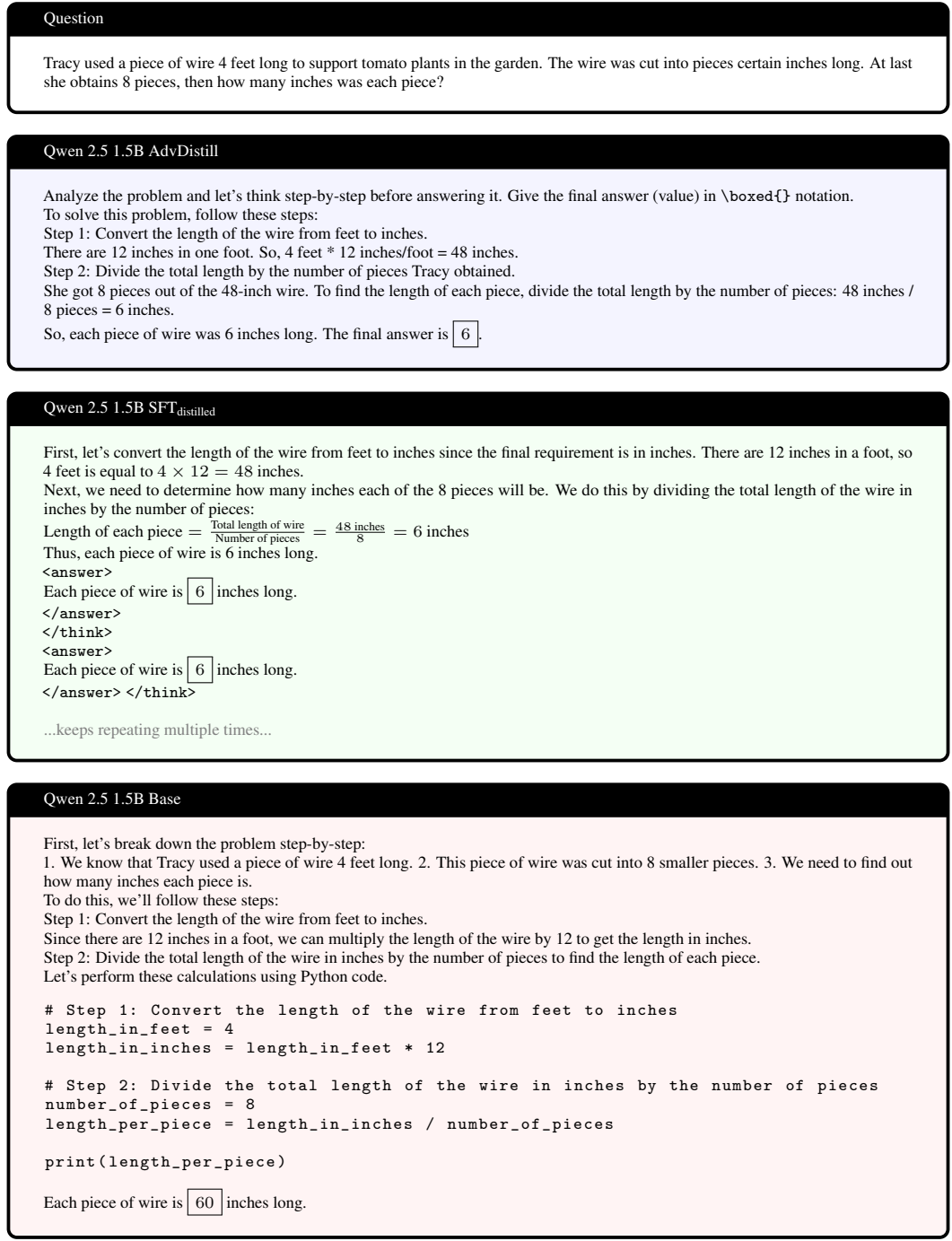

\scriptsize
\begin{tcolorbox}[title=Question, colframe=black, colback=white]
Tracy used a piece of wire 4 feet long to support tomato plants in the garden. The wire was cut into pieces certain inches long. At last she obtains 8 pieces, then how many inches was each piece?
\end{tcolorbox}

\vspace{0.1em}

% First model - Qwen 2.5 1.5B AdvDistill
\begin{tcolorbox}[title=Qwen 2.5 1.5B AdvDistill, colframe=black, colback=blue!4.5, width=\textwidth]
Analyze the problem and let's think step-by-step before answering it. Give the final answer (value) in \verb|\boxed{}| notation.

To solve this problem, follow these steps:

Step 1: Convert the length of the wire from feet to inches.\\
There are 12 inches in one foot. So, 4 feet * 12 inches/foot = 48 inches.

Step 2: Divide the total length by the number of pieces Tracy obtained.\\
She got 8 pieces out of the 48-inch wire. To find the length of each piece, divide the total length by the number of pieces: 48 inches / 8 pieces = 6 inches.

So, each piece of wire was 6 inches long. The final answer is $\boxed{6}$.
\end{tcolorbox}

\vspace{0.1em}

% Second model - Qwen 2.5 1.5B SFT
\begin{tcolorbox}[title=Qwen 2.5 1.5B SFT\textsubscript{distilled}, colframe=black, colback=green!4.5, width=\textwidth]
First, let's convert the length of the wire from feet to inches since the final requirement is in inches. There are 12 inches in a foot, so 4 feet is equal to $4 \times 12 = 48$ inches.

Next, we need to determine how many inches each of the 8 pieces will be. We do this by dividing the total length of the wire in inches by the number of pieces:

$\text{Length of each piece} = \frac{\text{Total length of wire}}{\text{Number of pieces}} = \frac{48 \text{ inches}}{8} = 6 \text{ inches}$

Thus, each piece of wire is 6 inches long.

\texttt{<answer>}\\
Each piece of wire is $\boxed{6}$ inches long.\\
\texttt{</answer>}

\texttt{</think>}\\
\texttt{<answer>}\\
Each piece of wire is $\boxed{6}$ inches long.\\
\texttt{</answer>}
\texttt{</think>}\\

\textcolor{gray}{...keeps repeating multiple times...}
\end{tcolorbox}

\vspace{0.1em}

% Third model - Qwen 2.5 1.5B Base
\begin{tcolorbox}[title=Qwen 2.5 1.5B Base, colframe=black, colback=red!4.5, width=\textwidth]
First, let's break down the problem step-by-step:

1. We know that Tracy used a piece of wire 4 feet long.
2. This piece of wire was cut into 8 smaller pieces.
3. We need to find out how many inches each piece is.

To do this, we'll follow these steps:

Step 1: Convert the length of the wire from feet to inches.\\
Since there are 12 inches in a foot, we can multiply the length of the wire by 12 to get the length in inches.

Step 2: Divide the total length of the wire in inches by the number of pieces to find the length of each piece.

Let's perform these calculations using Python code.
\begin{lstlisting}[basicstyle=\ttfamily\scriptsize]
# Step 1: Convert the length of the wire from feet to inches
length_in_feet = 4
length_in_inches = length_in_feet * 12

# Step 2: Divide the total length of the wire in inches by the number of pieces
number_of_pieces = 8
length_per_piece = length_in_inches / number_of_pieces

print(length_per_piece)
\end{lstlisting}

Each piece of wire is $\boxed{60}$ inches long.
\end{tcolorbox}

\caption{\textbf{Comparison of model outputs for the wire cutting problem.} The \texttt{AdvDistill} model provides a correct and concise solution with the answer $\boxed{6}$ inches, demonstrating improvement from the Base model despite not following the template with \texttt{<think>} tags. The SFT\textsubscript{distilled} model also correctly calculates 6 inches but exhibits a degenerate pattern of repeatedly outputting the same answer with tags. The Base model produces an incorrect answer of $\boxed{60}$ inches, making a computational error in the conversion or division process.}
\end{figure}

\newpage

\section{Model Configurations and Hyperparameters}

\begin{table}[ht]
\centering
\small
\begin{tabular}{l|c}
\toprule
\textbf{Parameter} & \textbf{Value} \\
\midrule
Batch Size & 32 \\
Number of Generations & 8 \\
Max Tokens per Response & 2048 \\
Max Context Length & 4096 \\
Temperature (Teacher) & 0.9 \\
Top-p Sampling & 1.0 \\
\bottomrule
\end{tabular}
\vspace{1em}
\caption{\textbf{Teacher model generation settings used to accumulate knowledge via multi-response prompting.}}
\label{tab:teacher_params}
\end{table}

\begin{table}[ht]
\centering
\small
\begin{tabular}{l|c}
\toprule
\textbf{Hyperparameter} & \textbf{Value} \\
\midrule
Batch Size & 1 (8 Responses) \\
Gradient Accumulation Steps & 16 \\
Optimizer & AdamW \\
Learning Rate (LR) & 5e-6 \\
Weight Decay & 0.01 \\
Epochs & 4 \\
Warmup Ratio & 0.05 \\
LR Scheduler & Cosine \\
Max Gradient Norm & 0.5 \\
Temperature (Student) & 0.5 \\
Lambda (Incorrect Response Loss) & 0.5 \\
Max Sequence Length & 2048 \\
Validation Steps & 400 \\
Precision & bfloat16 \\
\bottomrule
\end{tabular}
\vspace{1em}
\caption{\textbf{Student model (AdvDistill and SFT\textsubscript{distilled}) training hyperparameters used across all datasets.}}
\label{tab:student_params}
\end{table}

\newpage

\section{Model Optimizations and Effects}

\subsection{Degeneracy of Tokens}
\begin{table}[H]
\centering
\small
\begin{tabular}{l|cc|cc|cc|cc|cc}
\toprule
\multirow{2}{*}{\textbf{Model}} & \multicolumn{2}{c|}{\textbf{GSM-8k}} & \multicolumn{2}{c|}{\textbf{GSM-PLUS}} & \multicolumn{2}{c|}{\textbf{OPEN-S1}} & \multicolumn{2}{c|}{\textbf{OPEN-R1}} & \multicolumn{2}{c}{\textbf{MMLU-PRO}} \\
 & \checkmark & \ding{55} & \checkmark & \ding{55} & \checkmark & \ding{55} & \checkmark & \ding{55} & \checkmark & \ding{55} \\
\midrule
Qwen 2.5-1.5B Base & \textbf{0.027} & 0.019 & \textbf{0.021} & 0.007 & 0.356 & 0.224 & \textbf{0.494} & 0.092 & \textbf{2.884} & 0.115 \\
Qwen 2.5-3B Base & 0.004 & 0.006 & 0.020 & \textbf{0.014} & \textbf{0.700} & 0.132 & \textbf{0.567} & 0.063 & 0.061 & 0.007 \\
Qwen 2.5-7B Base & 0.000 & \textbf{0.023} & 0.001 & \textbf{0.002} & 0.339 & \textbf{0.142} & 0.227 & \textbf{0.063} & 0.059 & 0.005 \\
Qwen 2.5-1.5B SFT & 0.008 & \textbf{0.038} & 0.008 & \textbf{0.010} & \textbf{1.067} & 0.228 & 0.445 & \textbf{0.123} & 0.066 & \textbf{0.048} \\
Qwen 2.5-3B SFT & 0.001 & \textbf{0.018} & 0.000 & \textbf{0.008} & \textbf{0.686} & 0.151 & 0.021 & \textbf{0.011} & 0.021 & \textbf{0.011} \\
Qwen 2.5-1.5B AdvDistill & \textbf{0.040} & 0.038 & \textbf{0.049} & 0.017 & \textbf{0.957} & 0.044 & \textbf{0.244} & 0.019 & \textbf{0.244} & 0.019 \\
\bottomrule
\end{tabular}
\vspace{1em}
\caption{\textbf{Degenerate (recursive tokens) responses per 1000, separated into correct (\checkmark) and incorrect (\ding{55}) responses.} \textbf{Bold values} indicate the higher degeneracy rate for each dataset-condition pair.}
\label{tabledegenerate}
\end{table}

\subsection{Verbosity and Correctness}
\begin{table}[H]
\centering
\small
\begin{tabular}{l|cc|cc|cc|cc|cc}
\toprule
\multirow{2}{*}{\textbf{Model}} & \multicolumn{2}{c|}{\textbf{GSM-8k}} & \multicolumn{2}{c|}{\textbf{GSM-PLUS}} & \multicolumn{2}{c|}{\textbf{OPEN-S1}} & \multicolumn{2}{c|}{\textbf{OPEN-R1}} & \multicolumn{2}{c}{\textbf{MMLU-PRO}} \\
 & \checkmark & \ding{55} & \checkmark & \ding{55} & \checkmark & \ding{55} & \checkmark & \ding{55} & \checkmark & \ding{55} \\
\midrule
Qwen 2.5-1.5B Base & 199.9 & 214.6 & 213.4 & 240.6 & 345.9 & 461.0 & 284.1 & \textbf{413.0} & 662.9 & 717.9 \\
Qwen 2.5-3B Base & 147.9 & 171.9 & 197.5 & 232.4 & 333.3 & \textbf{451.3} & 312.9 & 423.9 & 297.1 & 309.9 \\
Qwen 2.5-7B Base & 142.7 & 169.9 & 192.0 & 216.3 & 339.5 & 456.8 & 314.2 & 411.2 & \textbf{507.7} & 450.0 \\
Qwen 2.5-1.5B SFT & 552.8 & 552.7 & 555.2 & 573.8 & 442.7 & 526.6 & 383.2 & 484.7 & 284.3 & \textbf{345.7} \\
Qwen 2.5-3B SFT & 140.0 & 175.8 & 163.2 & 214.6 & 341.8 & 445.1 & 282.2 & 329.6 & 282.2 & 329.6 \\
Qwen 2.5-1.5B AdvDistill & 178.4 & \textbf{343.0} & 195.9 & \textbf{350.3} & 332.8 & 370.4 & 326.0 & 361.0 & 326.0 & 361.0 \\
\bottomrule
\end{tabular}
\vspace{1em}
\caption{\textbf{Comparing response length with output accuracy (correct (\checkmark) /incorrect (\ding{55})) for each model and dataset}. \textbf{Bold values} indicate the largest difference between correct and incorrect response lengths for each dataset.}
\label{tableresponselength}
\end{table}

\begin{figure}[H]
    \centering
    \includegraphics[width=0.7\linewidth]{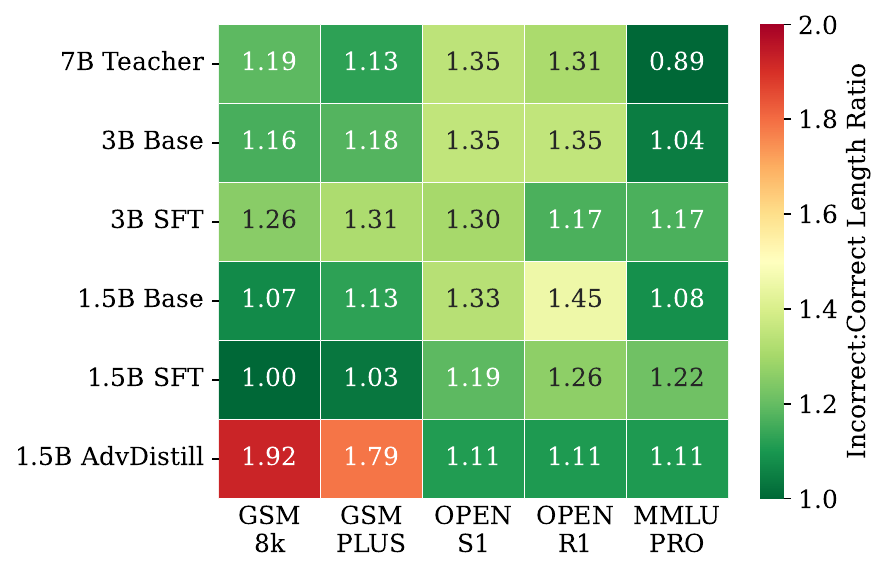}
    \caption{\textbf{Response verbosity (length) incorrect to correct ratio.}}
    \label{responselengthration}
\end{figure}

\newpage

\section{Evaluation Prompts}

\subsection{Mathematical Datasets (GSM-8K, GSM-PLUS)}

\begin{minipage}{\dimexpr\textwidth-2\fboxsep-2\fboxrule}
\begin{tcolorbox}[title=GSM8K \& GSM-Plus Evaluation Prompt]
\small{
\textcolor{brown}{// System Instruction}\\
You are a helpful AI assistant specializing in solving math problems. Please use XML-like <think> tag for reasoning, self-critique and verifying your thought process, <answer> tag with \verb|\boxed{}| for the final results.\\

\textcolor{brown}{// User Instruction}

Question: Janet's ducks lay 16 eggs per day. She eats 3 eggs for breakfast each morning and bakes muffins for her friends twice a week that require 6 eggs each time. She sells the remainder at the farmers market daily for \$2 per egg. How much in dollars does she make per week?

<think>\\
First, I'll calculate the total number of eggs laid by Janet's ducks in a week:

If 16 eggs are laid each day, then over 7 days: 16 × 7 = 112 eggs.\\
(Quick check: 7 × 10 = 70, 7 × 6 = 42, and 70 + 42 = 112 — checks out.)

Next, I'll verify Janet's weekly egg usage:\\
For breakfast: 3 eggs per day × 7 days = 21 eggs.\\
(3 × 7 = 21 — confirmed.)

For muffins: 6 eggs per batch × 2 batches per week = 12 eggs.\\
(6 × 2 = 12 — confirmed.)

Total consumed: 21 + 12 = 33 eggs.\\
(Double check: 20 + 10 = 30, plus 3 = 33 — correct.)\\
Now, I'll find how many eggs remain for selling:\\
Total laid (112) - Total used (33) = 79 eggs.\\
(Check: 112 - 30 = 82, then subtract 3 more = 79 — verified.)     \\
Finally, I'll compute weekly earnings:\\
79 eggs × \$2 each = \$158.\\
(Quick check: 80 × 2 = 160, minus 1 egg = 158 — matches.)\\
</think>

<answer>\\
Janet's ducks lay 16 eggs per day, so in a week they lay 16 × 7 = 112 eggs.\\
Janet uses 3 eggs for breakfast each day (21 eggs per week) and 6 eggs for muffins twice a week (12 eggs per week).\\
Total eggs used: 21 + 12 = 33 eggs per week\\
Eggs remaining to sell: 112 - 33 = 79 eggs per week\\
Income from selling eggs: 79 × \$2 = \$158 per week

The answer is \$ \verb|\boxed{158}|.\\
</answer>

// target question\\
Now solve this problem:\\
Question: ...
}
\end{tcolorbox}
\end{minipage}

\subsection{Complex Mathematics and Reasoning Datasets (OPEN-S1, OPEN-R1)}

\begin{minipage}{\dimexpr\textwidth-2\fboxsep-2\fboxrule}
\begin{tcolorbox}[title=OPEN-S1 \& OPEN-RS Evaluation Prompt]
\small{
\textcolor{brown}{// System Instruction}\\
You are a helpful AI assistant specializing in solving advanced mathematical problems. Please use <think> tags for reasoning, self-critique and verifying your thought process and <answer> tags with \\boxed{} notation for the final result.
\\
\textcolor{brown}{// User Instruction}\\
First, carefully analyze the problem and work through your solution in the <think></think> tags.\\
Then provide your final answer in the <answer></answer> tags. Finally, **make sure** the final numerical answer that you compute is inside \verb|\boxed{}| LaTex command.

- Verify: After writing, check whether you've given the final answer inside \verb|\boxed{}| LaTex command, if not, write it.

Now please solve this problem:\\
Question: ..
}
\end{tcolorbox}
\end{minipage}

\subsection{Multi-task and Knowledge Dataset (MMLU-PRO)}
\begin{minipage}{\dimexpr\textwidth-2\fboxsep-2\fboxrule}
\begin{tcolorbox}[title=MMLU-Pro Evaluation Prompt]
\small{
\textcolor{brown}{// System Instruction}\\
You are a helpful AI assistant specializing in advanced reasoning across multiple domains. Approach each problem methodically and carefully. Use <think> tags to work through your reasoning step-by-step, considering all aspects of the question, and use <answer> tags with \verb|\boxed{}| notation for your final answer letter.\\

\textcolor{brown}{// User Instruction}\\
Example 1:\\
Question: A marketing research firm contracts with clients to conduct a complete marketing research project from data collection, analysis, and reporting. It is a \rule{1cm}{0.15mm} firm.

Options:\\
A. Data collection firm.\\
B. Freelance agency.\\
C. Data analysis firm.\\
D. Survey firm.\\
E. Full-service agency.\\
F. Tabulation agency.\\
G. Marketing agency.\\
H. Research firm.\\
I. Consultant.\\
J. Field agency.

<think>\\
The question is asking about a type of marketing research firm that handles the entire process from data collection through analysis to reporting. Let me analyze each option:

A. Data collection firm - This would only handle the collection aspect, not the full process\\
B. Freelance agency - Too general, not specific to marketing research\\
C. Data analysis firm - This would focus only on analysis, not collection or reporting\\
D. Survey firm - This likely focuses mainly on surveys, not the complete process\\
E. Full-service agency - This indicates a firm that handles all aspects of the process\\
F. Tabulation agency - This focuses on organizing data, not the complete process\\
G. Marketing agency - Too general, could refer to an agency that does advertising or other marketing activities\\
H. Research firm - Too general, doesn't specify handling the complete process\\
I. Consultant - This suggests advisory services rather than conducting the full research\\
J. Field agency - This likely focuses on field research, not the complete process

Since the question specifically mentions a firm that conducts "a complete marketing research project from data collection, analysis, and reporting," option E "Full-service agency" best matches this description. A full-service agency would handle all aspects of the project as described.\\
</think>

<answer>\\
The marketing research firm that contracts with clients to conduct a complete marketing research project from data collection, analysis, and reporting would be a full-service agency.

\verb|\boxed{E}|\\
</answer>

Now please answer this question:\\
Question: ...
}
\end{tcolorbox}
\end{minipage}

%TC:endignore
\end{document}